\documentclass[journal,10pt,onecolumn]{IEEEtran}
\pdfoutput=1
\let\chapter\section

\usepackage{graphicx}
\usepackage{pifont}

\usepackage{booktabs}
\newcommand{\ra}[1]{\renewcommand{\arraystretch}{#1}}

\usepackage[caption=false,font=footnotesize]{subfig}
\usepackage[linesnumbered,ruled]{algorithm2e}
\usepackage{amsfonts}
\usepackage{amsmath}
\usepackage{amssymb,mathrsfs}
\usepackage{array}
\usepackage{amsthm}        
\usepackage{hyperref}
\usepackage{indentfirst}
\newtheorem{theorem}{Theorem}
\newtheorem{lemma}{Lemma}
\newtheorem{assumption}{Assumption}
\newtheorem{problem}{Problem}

\makeindex             

\usepackage[figuresright]{rotating}
\newcommand{\compresslist}{%
\setlength{\itemsep}{1pt}%
\setlength{\parskip}{0pt}%
\setlength{\parsep}{0pt}%
}

\allowdisplaybreaks[1]

\begin{document}
\title{Belief Space Planning Simplified: Trajectory-Optimized LQG (T-LQG)\\(Extended Report)}
\author{\IEEEauthorblockN{Mohammadhussein Rafieisakhaei, \and
Suman Chakravorty, and \and P. R. Kumar}\\
\IEEEauthorblockA{Texas A\&M University, College Station, Texas 77843, USA}\\
\IEEEauthorblockA{\{mrafieis, schakrav, prk\}@tamu.edu}}

\maketitle
\begin{abstract}
Planning under motion and observation uncertainties requires solution of a stochastic control problem in the space of feedback policies. In this paper, we reduce the general $ (n^2+n) $-dimensional belief space planning problem to an $ (n) $-dimensional problem by obtaining a Linear Quadratic Gaussian (LQG) design with the best nominal performance. Then, by taking the underlying trajectory of the LQG controller as the decision variable, we pose a coupled design of trajectory, estimator, and controller design through a Non-Linear Program (NLP) that can be solved by a general NLP solver. We prove that under a first-order approximation and a careful usage of the separation principle, our approximations are valid. We give an analysis on the existing major belief space planning methods and show that our algorithm has the lowest computational burden. Finally, we extend our solution to contain general state and control constraints. Our simulation results support our design.
\end{abstract}

\section{Introduction}\label{sec:Introduction}
Planning under process and measurement uncertainties is referred to as the belief space planning problem. In general, it requires the solution of Hamilton-Jacobi-Bellman (HJB) equations to obtain the optimal feedback policy \cite{Kumar-book-86,Bertsekas76}. The Linear Quadratic Gaussian (LQG) methodology provides the optimal estimator and controller for linear systems with Gaussian noises \cite{kumar2014control}. However, an LQG planner requires a nominal trajectory to begin with. An approach utilizing this methodology by decoupling the two procedures provides a sequential design of the trajectory and the LQG policy, either by providing a-priori trajectories and comparing the LQG performance over each one \cite{van2011lqg}, or by performing the decoupled procedure iteratively \cite{van2012motion,sun2016stochastic}. An approach to a coupled design of trajectory and the policy in non-linear systems utilizing the Extended Kalman Filter (EKF) is based on the heuristic assumption of Most-Likely Observations (MLO) during planning \cite{Platt10}. Another approach considers general belief distributions, either by reformulating the problem in belief space as a Partially Observed Markov Decision Process (POMDP) \cite{Sondik71} or by utilizing a Monte-Carlo representation of beliefs \cite{platt2013convex,rafieisakhaei2016feedbackICRA,rafieisakhaeiline}. An exact approach to solve POMDPs for continuous action and observation spaces requires continuous (uncountable) branching in the decision tree of beliefs, which leads to intractable computations. The state-of-the-art POMDP solvers are posed in countable decision trees, reducing the search space to discretized belief nodes. Similar to POMDP solvers, particle filter-based methods suffer from the curse of dimensionality due to the depletion of particles in the update step of the estimation.

In this paper, we overcome this hurdle by providing a coupled design of trajectory and the policy using the underlying trajectory as the optimization variable, and simplifying the belief space planning as an optimization problem aiming for estimation performance that can be solved by a general Non-Linear Programming (NLP) solver, which we refer to as the Trajectory-optimized LQG (T-LQG). Essentially, we use the separation principle and the structure of the LQG method to pose an optimization problem on the sequence of control actions over LQG polices, rather than optimizing over the general policy space. This method reduces the dimension of the underlying state in the optimization problem from $ n+n^2 $ (Gaussian belief dimension) to $ n $ (state dimension), breaking the computational burden of belief space planning problems. The computational complexity of our method is $ O(Kn^3) $, where $ K $ is the planning horizon and $ n $ is the state dimension, which is lower than any other belief space planning method.

We utilize the separation principle to separate the design of estimator and the controller. Then---assuming the existence of a linear controller to stabilize the system around a given nominal trajectory---we prove that under a first-order approximation, the stochastic control objective is dominated by the nominal part of the cost function. Moreover, over the given nominal trajectory, the nominal performance of the estimator is given by the dynamic Riccati equations independent from the observations and controller form. The original problem is reformulated and reduced to a deterministic problem over the state space by choosing the underlying nominal trajectory as the optimization variable, aiming for the best estimation performance over that trajectory. The key observation is: fixing the feedback policy to an LQG policy for a linearized system around a nominal trajectory offers the solution of an optimal estimation and control performance along that specific trajectory.

For a fixed linearization trajectory, LQG gives the best estimator and controller to track that nominal trajectory. Our method uses the nominal trajectory itself as an optimization variable in order to obtain the best trajectory, and, subsequently, the best estimator and controller to follow that trajectory. The central difference between this method and the MLO method of \cite{Platt10} is the utilization of the separation principle, which maintains a separate design of the controller to keep the state around the nominal trajectory. Otherwise, the state deviation from the nominal trajectory keeps growing intractably and the assumptions of the nominal trajectory (of control and subsequent state and observations) collapse, reducing the approach to a heuristic design as in \cite{Platt10}. On the contrary, in our approach, the controller keeps the state around the nominal trajectory, and therefore, the nominal estimation performance becomes valid. Also, using a high-dimensional controller such as a belief-LQR over an $ (n+n^2) $-dimensional space as in \cite{Platt10}---or \cite{van2012motion} and \cite{sun2016stochastic}---means avoiding a proper usage of the separation principle by coupling the controller design with the design of the estimator. Using the separation principle also enables us to pose the problem as a standard NLP, rather than using a dynamic programming mechanism (which involves tedious calculations in an $ (n+n^2) $-dimensional space to solve the coupled equations of the belief estimation and the controller design, as in \cite{van2012motion, sun2016stochastic}). 

Finally, whenever the accumulated error of linearization (or other errors) increases from a tolerable threshold during the execution, replanning occurs. This is also another merit of posing the planning problem as a standard NLP with low dimension: replanning for a long horizon becomes possible in online applications. Moreover, it enables the utilization of various optimization literature's highly optimized softwares and tools.

Unlike point-based POMDP solvers \cite{Pineau03,GShaniJPineau13,seiler2015online}, in T-LQG the time-horizon is a linear factor in the computational complexity, rather than a factor in the exponent---the curse of dimensionality (as in POMDPs and particle filter-based solvers) and the curse of history (as in POMDPs) disappears. This means T-LQG is capable of solving belief space problems on a considerably grander scale. Indeed, current point-based solvers are in short of solving problems in continuous action and observation spaces that are comparable to the problems that are dealt with in this paper. 

In the next section, we define the general stochastic control problem to be tackled. We provide our method in section \ref{sec:Our Method}, which includes the optimization problem, the execution and the replanning strategy. We provide a general method of dealing with non-convex path constraints in section \ref{sec:Non-Convex State Constraints}, and adapt our optimization to address that. In section \ref{sec:Theoretical Considerations}, we provide a theoretical analysis on the validity of our method and provide the sufficient conditions under which our solution is a valid approximation of the original problem. After providing an extensive theoretical comparison between the state-of-the-art belief space planning methods in section \ref{sec:Comparison of Methods}, we provide our simulation results for several situations in section \ref{sec:Simulation Results} and conclude afterwards. Note, a shorter version of this report has been submitted to WAFR, 2016 
\cite{Rafi2016WAFR}.
\section{General Problem}\label{sec:General Problem}
The general belief space planning problem is formulated as a stochastic control problem in the space of feedback policies. In this section, we define the basic elements of the problem, including system equations and belief dynamics.

\emph{System equations:} We denote the state, control and observation vectors by $\mathbf{x}\in \mathbb{X} \subset\mathbb{R}^{n_x}$, $\mathbf{u}\in \mathbb{U}\subset\mathbb{R}^{n_u}$, and $\mathbf{z}\in \mathbb{Z}\subset\mathbb{R}^{n_z}$, respectively. The motion $f:\mathbb{X}\times\mathbb{U}\times\mathbb{R}^{n_x}\rightarrow\mathbb{X}$, and observation $h:\mathbb{X}\rightarrow\mathbb{Z}$ processes are defined as:  
\begin{subequations}\label{eq:non-linear system equations}
\begin{alignat}{2}
\mathbf{x}_{t+1}&=f(\mathbf{x}_{t},\mathbf{u}_{t},\boldsymbol{\omega}_{t}), ~~&&\boldsymbol{\omega}_{t}\sim \mathcal{N}(\mathbf{0}, \boldsymbol{\Sigma}_{\boldsymbol{\omega}_{t}})\label{eq:linear-sys-general-app-2}\\
\mathbf{z}_{t}&=h(\mathbf{x}_{t},\boldsymbol{\nu}_{t}),~~ && \boldsymbol{\nu}_{t}\sim \mathcal{N}(\mathbf{0}, \boldsymbol{\Sigma}_{\boldsymbol{\nu}_{t}})
\end{alignat}
\end{subequations}
where $ \{\boldsymbol{\omega}_t\} $ and $ \{\boldsymbol{\nu}_t\} $ are zero mean independent, identically distributed (i.i.d.) mutually independent random sequences.

\textit{Belief (information state):} The conditional distribution of $ \mathbf{x}_t $ given the data history up to time $ t $, is called the belief $ \mathbf{b}_t:\mathbb{X}\times\mathbb{Z}^{t}\times\mathbb{U}^{t}\rightarrow\mathbb{R} $. It is defined as $ \mathbf{b}_{t}(\mathbf{x},\mathbf{z}_{0:t},\mathbf{u}_{0:t-1}, \mathbf{b}_0):=p_{\mathbf{X}_t|\mathbf{Z}_{0:t};\mathbf{U}_{0:t-1}}(\mathbf{x}|\mathbf{z}_{0:t};\mathbf{u}_{0:t-1};\mathbf{b}_0) $, and denoted by $ \mathbf{b}_t $ in this paper. We use a Gaussian representation of belief in belief space, $ \mathbb{B} $, \cite{Thrun2005}. The belief dynamics follow Bayesian update equations, summarized as a function $ \tau\!:\!\mathbb{B}\!\times\!\mathbb{U}\!\times\!\mathbb{Z}\rightarrow\mathbb{B} $, where $ \mathbf{b}_{t+1}=\tau(\mathbf{b}_{t},\mathbf{u}_t,\mathbf{z}_{t+1}) $ \cite{Kumar-book-86,Bertsekas07}.

\begin{problem}\label{problem:Stochastic Control Problem} \textup{\textbf{Stochastic Control Problem}} Given an initial belief state $ \mathbf{b}_{0} $, solve for the optimal policy as follows:
\begin{align}\label{problem eq:Stochastic Control Problem}
\nonumber \min_{\pi}~J^{\pi}:=\mathbb{E}[&\sum_{t=0}^{K-1}c_t^{\pi}(\mathbf{b}_t,\mathbf{u}_t)+c_K^{\pi}(\mathbf{b}_K)]
\\ s.t.~\mathbf{b}_{t+1}&=\tau(\mathbf{b}_{t},\mathbf{u}_t,\mathbf{z}_{t+1}),
\end{align}
where the optimization is over feasible policies, and $ \pi:=\{\pi_{0}, \cdots, \pi_{t}\} $ where $ \pi_{t} :\mathbb{B}\rightarrow \mathbb{U} $ specifies an action given the belief, $ \mathbf{u}_{t}=\pi_{t}(\mathbf{b}_{t}) $. Moreover, $ c^{\pi}_t(\cdot,\cdot):\mathbb{B}\times\mathbb{U}\rightarrow\mathbb{R} $ is the one-step cost function, $ c_K^{\pi}(\cdot):\mathbb{B}\rightarrow\mathbb{R} $  denotes the terminal cost, and the expectation is taken over all randomness.
\end{problem}

\section{Belief Space Planning Method: T-LQG}\label{sec:Our Method}
In this section, we provide the details of our design for the planning problem.

\subsection{Planning Problem}
Here, we provide the details of our design for the planning problem.

\textit{Parameterized Trajectories ($ p\textendash traj $)}: Using the noiseless equation of \eqref{eq:linear-sys-general-app-2} we parametrize the possible feasible propagated trajectories of the initial estimate, $ \hat{\mathbf{x}}_{0} $, given a set of unknown control inputs $ \{ \mathbf{u}^{p}_{t}\}_{t=0}^{K-1} $, as:
\begin{align*}
\mathbf{x}^{p}_{t+1}=f(\mathbf{x}^{p}_t, \mathbf{u}^{p}_t, 0),~~ 0\le t \le K\!-\!1
\end{align*}
where $ \mathbf{x}^{p}_{0} = \hat{\mathbf{x}}_{0} $. It is important to note that the trajectory $ \{\mathbf{x}^{p}_{t}\}_{t=0}^{K} $ changes by changing the underlying control inputs $ \{ \mathbf{u}^{p}_{t}\}_{t=0}^{K-1}  $. We will refer to the parametrized trajectory of $ \{\mathbf{x}^{p}_{t}\}_{t=0}^{K} $, $ \{ \mathbf{u}^{p}_{t}\}_{t=0}^{K-1}  $ as the $ p\textendash traj $.

\textit{Linearization of the system equations:} We linearize the non-linear motion and observation models of equation \eqref{eq:non-linear system equations} about the parametrized trajectory $ p\textendash traj $:
\begin{subequations}\label{eq:linearized system}
\begin{align}
\tilde{\mathbf{x}}_{t+1}&=\mathbf{A}^{p}_t(\mathbf{x}^{p}_{t},\mathbf{u}^{p}_t)\tilde{\mathbf{x}}_t + \mathbf{B}^{p}_t(\mathbf{x}^{p}_{t},\mathbf{u}^{p}_t)\tilde{\mathbf{u}}_t +\mathbf{G}^{p}_t(\mathbf{x}^{p}_{t},\mathbf{u}^{p}_t)\boldsymbol{\omega}_t\\
\tilde{\mathbf{z}}_{t}&=\mathbf{H}^{p}_t(\mathbf{x}^{p}_{t})\tilde{\mathbf{x}}_t+\mathbf{M}^{p}_t(\mathbf{x}^{p}_{t})\boldsymbol{\nu}_t,
\end{align}
\end{subequations}
where $ \mathbf{A}^{p}_t(\mathbf{x}^{p}_{t},\mathbf{u}^{p}_t)=\partial f(\mathbf{x},\mathbf{u},\boldsymbol{\omega})/\partial \mathbf{x}|_{ \mathbf{x}^{p}_{t}, \mathbf{u}^{p}_{t}, \mathbf{0} } $, $ \mathbf{B}^{p}_t(\mathbf{x}^{p}_{t},\mathbf{u}^{p}_t)=\partial f(\mathbf{x},\mathbf{u},\boldsymbol{\omega})/\partial \mathbf{u}|_{ \mathbf{x}^{p}_{t}, \mathbf{u}^{p}_{t}, \mathbf{0} } $, $ \mathbf{G}^{p}_t(\mathbf{x}^{p}_{t},\mathbf{u}^{p}_t)=\partial f(\mathbf{x},\mathbf{u}_t,\boldsymbol{\omega})/\partial \boldsymbol{\omega}|_{ \mathbf{x}^{p}_{t}, \mathbf{u}^{p}_{t}, \mathbf{0} } $, $ \mathbf{H}^{p}_t(\mathbf{x}^{p}_{t})=\partial h(\mathbf{x},\boldsymbol{\nu})/\partial \mathbf{x}|_{ \mathbf{x}^{p}_{t},\mathbf{0} } $ and $ \mathbf{M}^{p}_t(\mathbf{x}^{p}_{t})=\partial h(\mathbf{x},\boldsymbol{\nu})/\partial \boldsymbol{\nu}|_{ \mathbf{x}^{p}_{t},\mathbf{0} } $. Moreover $ \tilde{\mathbf{x}}_{t}:=\mathbf{x}_t-\mathbf{x}^{p}_{t} $, $ \tilde{\mathbf{u}}_{t}:=\mathbf{u}_t-\mathbf{u}^{p}_{t} $ and $ \tilde{\mathbf{z}}_{t}:=\mathbf{z}_t-\mathbf{z}^{p}_{t} $ define the state, control and observation errors, respectively, and $ \mathbf{z}^{p}_{t}=h(\mathbf{x}^{p}_{t}) $. Note that, as the control inputs change, the underlying trajectory they represent, $ \{\mathbf{u}^{p}_{t}\}_{t=0}^{K-1} $ and $ \{\mathbf{x}^{p}_{t}\}_{t=0}^{K} $, also change, and therefore Jacobian matrices change.

\textit{Expectation over policies:} In the cost function of problem \ref{problem:Stochastic Control Problem}, the expectation is over all types of policies in the feasible policy space. We assume the search is over linear feedback policies (LQG class), which is a valid assumption for locally controlling a linearized model around a nominal trajectory.

\textit{Separation principle:} It can be shown that the minimization of the stochastic cost function for a linear Gaussian system in terms of state error $ \mathbf{x}_t-\mathbf{x}^{p}_t $ is equivalent to two separate minimizations in terms of the estimation error $ \mathbf{x}_t-\hat{\mathbf{x}}_t $ and the controller error $ \hat{\mathbf{x}}_t-\mathbf{x}^{p}_t $, where $ \mathbf{x}_t-\mathbf{x}^{p}_t = \mathbf{x}_t-\hat{\mathbf{x}}_t+\hat{\mathbf{x}}_t-\mathbf{x}^{p}_t $ \cite{Bertsekas07}. As a result, the design of a stochastic controller with partially observed states reformulates to two separate designs of an estimator and a fully observed controller. In the LQG case, the estimator is a Kalman filter and the controller is an LQR controller.

\textit{Estimation cost:} The optimization cost function, $ J $, is a quadratic cost:
\begin{align}
J:=&\mathbb{E}[\sum_{t=1}^{K}\check{\mathbf{x}}^{T}_t\mathbf{W}^{x}_{t}\check{\mathbf{x}}_t+\tilde{\mathbf{u}}^{T}_{t-1}\mathbf{W}^{u}_{t}\tilde{\mathbf{u}}_{t-1}],\label{eq:Estimator cost}
\end{align}
where $ \check{\mathbf{x}}_{t}:=\mathbf{x}_t-\hat{\mathbf{x}}_{t} $ is the estimator error (KF error), and $ \mathbf{W}^{x}_{t}, \mathbf{W}^{u}_{t}\succeq 0$ are two positive-definite weight matrices. Moreover, we assume $ \mathbf{W}^{x}_{t} $ is symmetric. Therefore, these matrices have square root. Note, this cost function can be rewritten as a function of the belief trajectory, which is shown in Appendix \ref{appdx:Expressing the Cost Function as a Function of The Belief}. However, as we show in section \ref{sec:Theoretical Considerations}, under a first-order approximation, $ J $ is dominated by the part of the cost function that depends only on the underlying nominal trajectory, which is the $ p\textendash traj $ in this case. Thus, we approximate $ J $ with $ J^p $ as follows:
\begin{align}
J^{p}:=&\sum_{t=1}^{K}(\mathbb{E}[\tilde{\mathbf{x}}^{T}_t\mathbf{W}^{x}_{t}\tilde{\mathbf{x}}_t]+({\mathbf{u}}^{p}_{t-1})^{T}\mathbf{W}^{u}_{t}{\mathbf{u}}^{p}_{t-1}),\label{eq:LQG cost}
\end{align}
where the first term is a function of the nominal belief which is discussed next.

\textit{Nominal evolution of covariance:} Define $ \mathbf{P}^{+}_{\mathbf{b}^{p}_t}(\mathbf{x}^{p}_{0:t},\mathbf{u}^{p}_{0:t-1}):=\mathbb{E}[(\mathbf{x}_t-\mathbf{x}^{p}_{t})(\mathbf{x}_t-\mathbf{x}^{p}_{t})^T] $ to be the covariance of the nominal belief along the $ p\textendash traj $. Therefore, the first term of the cost function \eqref{eq:LQG cost}, can be rewritten as $ \sum_{t=1}^{K}\mathrm{tr}(\mathbf{W}_t \mathbf{P}^{+}_{\mathbf{b}^{p}_t}\mathbf{W}^T_t) $, where $ \mathbf{W}^{x}_{t}=\mathbf{W}^{T}_{t}\mathbf{W}_{t} $ is the Cholesky decomposition of $ \mathbf{W}^{x}_{t} $ (see Appendix \ref{appdx:Expressing the Cost Function as a Function of The Belief}). For a presumptive parametrized trajectory $ p\textendash traj $, the evolution of $ \mathbf{P}^{+}_{\mathbf{b}^{p}_t} $ is given by the recursive Riccati equations independent from the observations but as a function of the trajectory itself. Note, this also provides the nominal performance of the estimation along that trajectory. The dependencies on the $ p\textendash traj $ inside the parenthesis is dropped for simplicity. The evolution of $ \mathbf{P}^{+}_{\mathbf{b}^{p}_t} $ is given as follows:
\begin{align*}
\mathbf{P}^{-}_{\mathbf{b}^{p}_t}(\mathbf{x}^{p}_{0:t},\mathbf{u}^{p}_{0:t-1})&:=\mathbf{A}^{p}_t(\mathbf{x}^{p}_{t},\mathbf{u}^{p}_t)\mathbf{P}^{+}_{\mathbf{b}^{p}_{t-1}}(\mathbf{x}^{p}_{0:t-1},\mathbf{u}^{p}_{0:t-2})(\mathbf{A}^{p}_t(\mathbf{x}^{p}_{t},\mathbf{u}^{p}_t))^T+\mathbf{G}^{p}_t(\mathbf{x}^{p}_{t},\mathbf{u}^{p}_t)\boldsymbol{\Sigma}_{\boldsymbol{\omega}_{t}}(\mathbf{G}^{p}_t(\mathbf{x}^{p}_{t},\mathbf{u}^{p}_t))^T\\
\mathbf{S}_{\mathbf{b}^{p}_t}(\mathbf{x}^{p}_{0:t},\mathbf{u}^{p}_{0:t-1})&:=\mathbf{H}^{p}_t(\mathbf{x}^{p}_{t})\mathbf{P}^{-}_{\mathbf{b}^{p}_t}(\mathbf{x}^{p}_{0:t},\mathbf{u}^{p}_{0:t-1})(\mathbf{H}^{p}_t(\mathbf{x}^{p}_{t}))^T+\mathbf{M}^{p}_t(\mathbf{x}^{p}_{t})\boldsymbol{\Sigma}_{\boldsymbol{\nu}_{t}}(\mathbf{M}^{p}_t(\mathbf{x}^{p}_{t}))^T\\
\mathbf{K}_{\mathbf{b}^{p}_t}(\mathbf{x}^{p}_{0:t},\mathbf{u}^{p}_{0:t-1})&:=\mathbf{P}^{-}_{\mathbf{b}^{p}_t}(\mathbf{x}^{p}_{0:t},\mathbf{u}^{p}_{0:t-1})(\mathbf{H}^{p}_t(\mathbf{x}^{p}_{t}))^T(\mathbf{S}_{\mathbf{b}^{p}_t}(\mathbf{x}^{p}_{0:t},\mathbf{u}^{p}_{0:t-1}))^{-1}\\
\mathbf{P}^{+}_{\mathbf{b}^{p}_t}(\mathbf{x}^{p}_{0:t},\mathbf{u}^{p}_{0:t-1})&=(\mathbf{I}-\mathbf{K}_{\mathbf{b}^{p}_t}(\mathbf{x}^{p}_{0:t},\mathbf{u}^{p}_{0:t-1})\mathbf{H}^{p}_t(\mathbf{x}^{p}_{t}))\mathbf{P}^{-}_{\mathbf{b}^{p}_t}(\mathbf{x}^{p}_{0:t},\mathbf{u}^{p}_{0:t-1})
\end{align*}
which is solvable with the initial condition $ \mathbf{P}^{+}_{\mathbf{b}^{p}_0}(\mathbf{x}^{p}_{0})=\boldsymbol{\Sigma}_{\mathbf{x}_{0}} $.

\begin{problem}\label{problem:Planning Problem}\textup{\textbf{Planning Problem}} Given an initial belief $ \mathbf{b}_{0}\in\mathbb{B} $, a goal region represented as an $ \ell_2 $-norm ball, $ \mathbf{B}_{r_g}(\mathbf{x}_g) $, of radius $ r_g $ around a goal state $ \mathbf{x}_{g}\in\mathbb{X} $, and a planning horizon of $ K>0 $, we define the following problem:
\begin{subequations}
\begin{align}
\nonumber \min_{\mathbf{u}^{p}_{0:K-1}}\sum\limits_{t=1}^{K}[&\mathrm{tr}(\mathbf{W}_t \mathbf{P}^{+}_{\mathbf{b}^{p}_t}\mathbf{W}^T_t)+({\mathbf{u}}^{p}_{t-1})^{T}\mathbf{W}^{u}_{t}{\mathbf{u}}^{p}_{t-1}]
\\s.t.~~\mathbf{P}^{-}_{\mathbf{b}^{p}_t}&=\mathbf{A}^{p}_t\mathbf{P}^{+}_{\mathbf{b}^{p}_{t-1}}(\mathbf{A}^{p}_t)^T+\mathbf{G}^{p}_t\boldsymbol{\Sigma}_{\boldsymbol{\omega}_{t}}(\mathbf{G}_t)^T\label{eq:riccati 1}
\\\mathbf{S}_{\mathbf{b}^{p}_t}&=\mathbf{H}^{p}_t\mathbf{P}^{-}_{\mathbf{b}^{p}_t}(\mathbf{H}^{p}_t)^T+\mathbf{M}^{p}_t\boldsymbol{\Sigma}_{\boldsymbol{\nu}_{t}}(\mathbf{M}^{p}_t)^T\label{eq:riccati 2}
\\\mathbf{K}_{\mathbf{b}^{p}_t}&=\mathbf{P}^{-}_{\mathbf{b}^{p}_t}(\mathbf{H}^{p}_t)^T\mathbf{S}_{\mathbf{b}^{p}_t}^{-1}\label{eq:riccati 3}
\\\mathbf{P}^{+}_{\mathbf{b}^{p}_t}&=(\mathbf{I}-\mathbf{K}_{\mathbf{b}^{p}_t}\mathbf{H}^{p}_t)\mathbf{P}^{-}_{\mathbf{b}^{p}_t}\label{eq:riccati 4}
\\\mathbf{P}^{+}_{\mathbf{b}^{p}_0}&=\boldsymbol{\Sigma}_{\mathbf{x}_{0}}\label{eq:initial mean}
\\\mathbf{x}^{p}_{0} &= \mathbb{E}_{\mathbf{x}}[\mathbf{b}_{0}(\mathbf{x})]\label{eq:initial cov}
\\\mathbf{x}^{p}_{t+1}&=f(\mathbf{x}^{p}_t, \mathbf{u}^{p}_t, 0) ~0\!\le\! t\! \le\! K\!-\!1\label{eq:state propagation}
\\|\!|\mathbf{x}^{p}_{K}-\mathbf{x}_g|\!|_{2}&<r_g\label{eq:terminal constraint}
\\|\!|\mathbf{u}^{p}_{t}|\!|_{2}&\le r_u, ~1\!\le\! t\! \le\! K,\label{eq:control contraint}
\end{align}
\end{subequations}
where equations \eqref{eq:riccati 1}-\eqref{eq:riccati 4} are regarded as one constraint at each time step, and are used to calculate first term of the objective at that time step, equations \eqref{eq:initial mean} and \eqref{eq:initial cov} represent the initial conditions, equation \eqref{eq:state propagation} defines the state propagation (and relates the optimization variables to the state trajectory), equation \eqref{eq:terminal constraint} constrains the terminal state within an $ \ell_2 $-norm ball of radius $ r_g>0 $ around the goal state, equation \eqref{eq:control contraint} accounts for the saturation constraints for $ r_u>0 $. Moreover, the first term of the objective aims for minimizing the estimation uncertainty, whereas the second term penalizes the control effort. Note, this problem should be viewed as an optimization in the space of control actions with the control actions as the variables and every other variable, such as the covariances, as a function of those controls.
\end{problem}

\textit{Optimized Trajectory, $ o\textendash traj $:} We will denote the resulting optimized $ p\textendash traj $ of problem \ref{problem:Planning Problem} with $ \{\mathbf{x}^{o}_{t}\}_{t=0}^{K} $, $ \{ \mathbf{u}^{o}_{t}\}_{t=0}^{K-1}  $ and refer to it as the $ o\textendash traj $.

\textit{Feedback control:} The resulting trajectory from the optimization problem is optimized in terms of estimation performance. Now, using the separation principle, the LQR controller is designed to follow the $ o\textendash traj $. Therefore, the LQR cost is designed for the controller error $ \hat{\mathbf{x}}_t-\mathbf{x}^{o}_t $. The resulting control policy is a linear feedback policy, and the evolution of $ \hat{\mathbf{x}}_t $ is obtained from the KF equations using the actual observations in the execution. The details of these equations, which are common to any LQG problem, are discussed next.

\textit{Linearization of system equations:} Note that the linearization of the system equations still follow the same procedure as explained before in equations \eqref{eq:linearized system}. This is because, the $ o\textendash traj $ is a $ p\textendash traj $, as well. However, for simplicity, we would denote the Jacobian matrices and every other variable associated with the $ o\textendash traj $ with an $ o $ superscript. Thus, for the rest of the paper, the system equations are linearized around the $ o\textendash traj $, with $ \mathbf{A}^{o}_t=\partial f(\mathbf{x},\mathbf{u},\boldsymbol{\omega})/\partial \mathbf{x}|_{ \mathbf{x}^{o}_{t}, \mathbf{u}^{o}_{t}, \mathbf{0} } $, $ \mathbf{B}^{o}_t=\partial f(\mathbf{x},\mathbf{u},\boldsymbol{\omega})/\partial \mathbf{u}|_{ \mathbf{x}^{o}_{t}, \mathbf{u}^{o}_{t}, \mathbf{0} } $, $ \mathbf{G}^{o}_t=\partial f(\mathbf{x},\mathbf{u}_t,\boldsymbol{\omega})/\partial \boldsymbol{\omega}|_{ \mathbf{x}^{o}_{t}, \mathbf{u}^{o}_{t}, \mathbf{0} } $, $ \mathbf{H}^{o}_t=\partial h(\mathbf{x},\boldsymbol{\nu})/\partial \mathbf{x}|_{ \mathbf{x}^{o}_{t},\mathbf{0} } $ and $ \mathbf{M}^{o}_t=\partial h(\mathbf{x},\boldsymbol{\nu})/\partial \boldsymbol{\nu}|_{ \mathbf{x}^{o}_{t},\mathbf{0} } $. Moreover, define $ \mathbf{f}^{o}_{t}:=f(\mathbf{x}^{o}_{t},\mathbf{u}^{o}_{t}, \mathbf{0})-\mathbf{A}^{o}_t\mathbf{x}^{o}_t-\mathbf{B}^{o}_t\mathbf{u}^{o}_t $, $ \mathbf{h}^{o}_t:=h(\mathbf{x},\mathbf{0})- \mathbf{H}^{o}_t\mathbf{x}^{o}_t $.

\textit{LQR cost function:} The feedback controller cost function, $ J^{f} $ (with the superscript $ f $ accounting for the feedback controller), is a common quadratic cost, as follows:
\begin{align*}
J^{f}:=&\sum_{t=1}^{K}[(\hat{\mathbf{x}}_t-\mathbf{x}^{o}_t)^T\mathbf{W}^{x}_{t}(\hat{\mathbf{x}}_t-\mathbf{x}^{o}_t)+(\tilde{\mathbf{u}}^{o}_{t-1})^{T}\mathbf{W}^{u}_{t}\tilde{\mathbf{u}}^{o}_{t-1}],
\end{align*}
where $ \tilde{\mathbf{u}}^{o}_t = \mathbf{u}_t-\mathbf{u}^{o}_t $ and $ \mathbf{W}^{x}_{t}, \mathbf{W}^{u}_{t} $ are as defined before.

\textit{Control policy:} The resulting control policy is proved to be a linear feedback policy as follows:
\begin{align*}
\tilde{\mathbf{u}}^{o}_{t}=-\mathbf{L}^{o}_{t}(\hat{\mathbf{x}}_t-\mathbf{x}^{o}_t),
\end{align*}
where the linear feedback gain $ \mathbf{L}^{o}_{t} $ is given as follows:
\begin{align*}
\mathbf{L}^{o}_{t} = (\mathbf{W}^{u}_{t}+(\mathbf{B}^{o}_{t})^T\mathbf{P}^{f}_{t}\mathbf{B}^{o}_{t})^{-1}(\mathbf{B}^{o}_{t})^T\mathbf{P}^{f}_{t}\mathbf{A}^{o}_{t},
\end{align*}
and the matrix $ \mathbf{P}^{f}_{t} $ is the result of backward iterations of the dynamic Riccati equation as follows:
\begin{align*}
\mathbf{P}^{f}_{t-1} = (\mathbf{A}^{o}_{t})^T\mathbf{P}^{f}_{t}\mathbf{A}^{o}_{t}-(\mathbf{A}^{o}_{t})^T\mathbf{P}^{f}_{t}\mathbf{B}^{o}_{t}(\mathbf{W}^{u}_{t}+(\mathbf{B}^{o}_{t})^T\mathbf{P}^{f}_{t}\mathbf{B}^{o}_{t})^{-1}(\mathbf{B}^{o}_{t})^T\mathbf{P}^{f}_{t}\mathbf{A}^{o}_{t}+\mathbf{W}^{x}_{t},
\end{align*}
which is solvable with a terminal condition $ \mathbf{P}^{f}_{K}=\mathbf{W}^{x}_{t} $.


\textit{Estimation Trajectory, $ e\textendash traj $:} The estimate of the state during the execution can be obtained using a Kalman Filter. This filtered trajectory is denoted by $ \{\hat{\mathbf{x}}_{t}\}_{t=0}^{K} $, and called the $ e\textendash traj $. The equations governing its dynamics and relations to the corresponding control actions are derived next.

\textit{Estimation with KF:} coupled with the LQR controller, in an LQG control strategy lies an Kalman filter to obtain the updated posterior distribution, whose mean update equations are as follows:
\begin{align}
\hat{\mathbf{x}}_{t+1} = (\mathbf{I}-\mathbf{K}^{o}_{t+1}\mathbf{H}^{o}_{t+1})\mathbf{f}^{o}_{t}-\mathbf{K}^{o}_{t+1}\mathbf{h}^{o}_{t+1}+\mathbf{A}^{o}_t\hat{\mathbf{x}}_t + \mathbf{B}^{o}_t\mathbf{u}_t + \mathbf{K}^{o}_{t+1}(\mathbf{z}_{t+1}-\mathbf{H}^{o}_{t+1}(\mathbf{A}^{o}_t\hat{\mathbf{x}}_t + \mathbf{B}^{o}_t\mathbf{u}_t))\label{eq:KF mean update}
\end{align}
with $ \hat{\mathbf{x}}_{0} = \mathbb{E}(\mathbf{x}_0)  $ (see Appendix \ref{appdx:Kalman Filtering Mean Update} for derivation). Moreover, $ \mathbf{K}^{o}_{t} $ is the Kalman gain whose equation is as follows:
\begin{align*}
\mathbf{K}^{o}_{t}=\textbf{A}^{o}_{t}\textbf{P}^{o}_{t}(\mathbf{H}^{o}_{t})^T(\mathbf{H}^{o}_{t}\textbf{P}^{o}_{t}(\mathbf{H}^{o}_{t})^T+\mathbf{M}^{o}_{t}\boldsymbol{\Sigma}_{\boldsymbol{\nu}_{t}}(\mathbf{M}^{o}_{t})^T)^{-1},
\end{align*}
and the matrix $ \textbf{P}^{o}_{t}:=\mathbf{E}[(\mathbf{x}_{t}-\mathbf{x}^{o}_{t})(\mathbf{x}_{t}-\mathbf{x}^{o}_{t})^{T}] $ is the covariance of the state estimation, whose governing equations are the forward Riccati equations as follows:
\begin{align}
\textbf{P}^{o}_{t+1} = \textbf{A}^{o}_{t}(\textbf{P}^{o}_{t} - \textbf{P}^{o}_{t}(\mathbf{H}^{o}_{t})^T(\mathbf{H}^{o}_{t}\textbf{P}^{o}_{t}(\mathbf{H}^{o}_{t})^T+\mathbf{M}^{o}_{t}\boldsymbol{\Sigma}_{\boldsymbol{\nu}_{t}}(\mathbf{M}^{o}_{t})^T)^{-1}\textbf{H}^{o}_{t}\textbf{P}^{o}_{t}+\mathbf{G}^{o}_{t}\boldsymbol{\Sigma}_{\boldsymbol{\omega}_{t}}(\mathbf{G}^{o}_{t})^T)(\textbf{A}^{o}_{t})^{T},
\end{align}
which is solvable with the initial condition of $ \textbf{P}^{o}_{0}:=\mathbf{E}[(\mathbf{x}_{0}-\mathbf{x}^{o}_{0})(\mathbf{x}_{0}-\mathbf{x}^{o}_{0})^{T}] $.

\subsection{Replanning During Execution}

\textit{Replanning during execution:} In a stochastic system, even with a closed loop control strategy, after a finite number of execution steps, the estimate deviates from the planned trajectory. This happens due to the accumulation of errors resulting from the un-modeled dynamics, noise, non-linearities, and unpredicted forces. In such a situation, the planned policy becomes irrelevant and a new policy is needed to drive the agent toward the predefined goals. In order to overcome the problem, we track the nominal belief. Then, based on the Kullback-Leibler divergence between the nominal and true belief, we define a symmetric distance (average of unsymmetrical KL-divergences). Once such a distance is greater than a predefined threshold  $ d_{th}>0 $, a deviation is detected, the planning module is initialized with the current belief, and all planning procedures are performed again. The details are discussed next.
 
\textit{Kullback-Leibler (KL) divergence:} The KL divergence itself is not a symmetric distance function, however, a symmetric distance can be easily derived from that. If $ D_{KL}(Q_1 \parallel Q_2) $ denotes the KL divergence of $ Q_1 $ and $ Q_2 $, where the latter are two probability distributions, then $ d(Q_1, Q_2):=(D_{KL}(Q_1 \parallel Q_2)+D_{KL}(Q_2 \parallel Q_1))/2 $ denotes a distance between $ Q_1 $ and $ Q_2 $, where
\begin{align*}
D_{KL}(Q_1 \parallel Q_2) = \int_{-\infty}^{\infty} q_1(\mathbf{x})\log(\frac{q_1(\mathbf{x})}{q_2(\mathbf{x})})d\mathbf{x},
\end{align*}
with $ q_1(\mathbf{x}) $ and $ q_2(\mathbf{x}) $ denoting the densities of $ Q_1 $ and $ Q_2 $.

\textit{Detection of a deviation:} At any time instance, the estimation model reports the current belief as $ \mathbf{b}_{t} = \mathcal{N}(\hat{\mathbf{x}}_{t}, \mathbf{P}^{o}_{t}) $, whereas the nominal belief at $ t $ is $ \mathbf{b}^{o}_{t} = \mathcal{N}(\mathbf{x}^{o}_{t}, \mathbf{P}^{o}_{t}) $. Therefore, using the KL divergence formula for multivariate Gaussian distributions \cite{duchi2007derivations}, the distance between $ \mathbf{b}_t $ and $ \mathbf{b}^{o}_t $ is as follows:
\begin{align}
\nonumber d(\mathbf{b}_t, \mathbf{b}^{o}_t) =& \frac{1}{4}[\log\frac{|\mathbf{P}^{o}_{t}|}{|\mathbf{P}^{o}_{t}|}-n_x+\mathrm{tr}((\mathbf{P}^{o}_{t})^{\!-\!1}\mathbf{P}^{o}_{t})+(\mathbf{x}^{o}_{t}-\hat{\mathbf{x}}_{t})^{T}(\mathbf{P}^{o}_{t})^{\!-\!1}(\mathbf{x}^{o}_{t}-\hat{\mathbf{x}}_{t})]
\\\nonumber&+\frac{1}{4}[\log\frac{|\mathbf{P}^{o}_{t}|}{|\mathbf{P}^{o}_{t}|}-n_x+\mathrm{tr}((\mathbf{P}^{o}_{t})^{\!-\!1}\mathbf{P}^{o}_{t})+(\hat{\mathbf{x}}_{t}-\mathbf{x}^{o}_{t})^{T}(\mathbf{P}^{o}_{t})^{\!-\!1}(\hat{\mathbf{x}}_{t}-\mathbf{x}^{o}_{t})]
\\=&\frac{1}{2}[n_x(n_x\!-\!1)\!+\!(\hat{\mathbf{x}}_{t}\!-\!\mathbf{x}^{o}_{t})^{T}(\mathbf{P}^{o}_{t})^{\!-\!1}(\hat{\mathbf{x}}_{t}\!-\!\mathbf{x}^{o}_{t})],
\end{align}
where $ |\mathbf{P}| $ denotes the determinant of the matrix $ \mathbf{P} $. Thus a deviation is detected if $ d(\mathbf{b}_t, \mathbf{b}^{o}_t)>d_{th} $.

\section{Non-Convex State Constraints}\label{sec:Non-Convex State Constraints}
Non-convex state constraints are handled with barrier functions. 

\textit{Polygonal obstacles approximated by ellipsoids:} Given a set of vertices that constitute a polygonal obstacle, we find the Minimum Volume Enclosing Ellipsoid (MVEE) and obtain its parameters \cite{moshtagh2005minimum}. Particularly, for the $ i $th obstacle, the barrier function includes a Gaussian-like function, where the argument of the exponential is the MVEE, which can be disambiguated with its center $ \mathbf{c}^i $ and a positive definite matrix $ \mathbf{E}^i $ that determines the rotation and axes of the ellipsoid. Moreover, we add several number of inverse functions that tend to infinity along the major and minor axes of the ellipsoid. So, the overall function acts as a barrier to prevent the trajectory from entering the region enclosed by the ellipsoid. Note that for non-polygonal obstacles, one can find the MVEE, and the algorithm works independently from this fact. Therefore, given the ellipsoid parameters $ \mathcal{C} := (\mathbf{c}^1, \mathbf{c}^2,\cdots, \mathbf{c}^{n_b})\in\mathbb{R}^{n_x\times n_b} $ and $ \mathcal{E} := (\mathbf{E}^1,\cdots, \mathbf{E}^{n_b})\in\mathbb{R}^{n^2_x\times n_b} $, the Obstacle Barrier Function (OBF) is constructed as follows:
\begin{align*} 
\nonumber \Phi^{(\mathcal{E},\mathcal{C})}(\mathbf{x})\!\!:=&\!\sum_{i=1}^{n_b}\Big[M_1\exp(-[(\mathbf{x}-\mathbf{c}^i)^T\mathbf{E}^i(\mathbf{x}-\mathbf{c}^i)]^q)
+M_2\!\!\!\!\!\!\!\sum\limits_{\theta=0:\epsilon_m:1}\!\!\!\!\!\!(\!|\!|\mathbf{x}\!\!-\!(\theta \zeta^{i,1}\!+\!(1\!\!-\!\theta) \zeta^{i,2})|\!|^{-\!2}_{2}\!\!+\!|\!|\mathbf{x}\!-\!(\theta \xi^{i,1}\!+\!(1\!\!-\!\theta) \xi^{i,2})|\!|^{-\!2}_{2})\!\Big],
\end{align*}
where $ \epsilon_m = 1/m, m\in \mathbb{Z}^{+} $, $ M_1, M_2\ge 0 $, $ q\in\mathbb{Z}^{+} $, and $ \zeta^{i,1} $, $ \zeta^{i,2} $ and $ \xi^{i,1} $, $ \xi^{i,2} $ are the endpoints of the major and minor axes of the ellipsoid, respectively. Therefore, the second term in the sum places inverse function whose values tend to infinity along the axes of the ellipsoid at points formed by convex combination of the two endpoints of each axis. As $ \epsilon_m $ tends to zero, the entire axes of the ellipsoid become infinite, and, therefore, act as a barrier to any continuous trajectory of states. One can think of putting more infinity points inside the ellipsoid by forming the convex combination of the existing infinity points. Moreover, the first summand determines the territory of the ellipsoid and imposes an outwards gradient around the ellipsoid, acting as a penalty function pushing the trajectory out of the banned region. Hence, we define the cost of avoiding obstacles as:
\begin{align}\label{eq:cost obs integral}
{\mathrm{cost}}_{obst}(\mathbf{x}_{t1},\mathbf{x}_{t2}):=\int_{\mathbf{x}_{t1}}^{\mathbf{x}_{t2}}\Phi^{(\mathcal{E},\mathcal{C})}(\mathbf{x}')d\mathbf{x}',
\end{align}
which is the line integral of the OBF between two given points of the trajectory $ \mathbf{x}_{t1} $ and $ \mathbf{x}_{t2} $. Therefore, the addition of this cost to the optimization objective, ensures the solver minimizes this cost and keeps the trajectory out of banned regions. However, for implementation purposes, the integral in equation \eqref{eq:cost obs integral} is approximated by a finite Riemann sum consisting of fewer points between $ \mathbf{x}_{t1} $ and $ \mathbf{x}_{t2} $. 
Using this equation, we add the running obstacle cost of $ {\mathrm{cost}}_{obst}(\mathbf{x}_{t-1},\mathbf{x}_{t}) $ to the optimization objective and use the modified optimization problem to obtain locally optimal solutions in the inter-obstacle feasible space using gradient descent methods \cite{boyd2004convex}.

\section{Theoretical Considerations}\label{sec:Theoretical Considerations}
In this section, we derive the conditions upon which our planning strategy is an acceptable approximation of the original problem. We perform a first-order approximation on the propagated errors of the state, control, observation, belief, and cost function around a nominal trajectory that stems from a nominal trajectory of the control actions. We show that under some sufficiency conditions, our planning strategy provides an acceptable approximation to the original problem. Note all proofs of the lemmas and the theorem are provided in Appendix \ref{appdx:Proofs of Lemmas and Theorems}.

\textit{Linearization of process and observation models:} Assuming there exists a nominal trajectory of control actions, $ \{\mathbf{u}^{p}_t\}_{t=0}^{K-1} $, then there exists a corresponding nominal trajectory of states, $ \{\mathbf{x}^{p}_t\}_{t=0}^{K} $ and observations $ \{\mathbf{z}^{p}_t\}_{t=0}^{K-1} $, where $ \mathbf{x}^{p}_{t+1}=f(\mathbf{x}^{p}_{t},\mathbf{u}^{p}_{t}, \mathbf{0}) $, $ \mathbf{z}^{p}_t=h(\mathbf{x}^{p}_t) $, and the state, control, and observation error vectors are $ \tilde{\mathbf{u}}_{t}=\mathbf{u}_t - \mathbf{u}^{p}_t $, $ \tilde{\mathbf{x}}_{t}=\mathbf{x}_t - \mathbf{x}^{p}_t $ and $ \tilde{\mathbf{z}}_t =\mathbf{z}_t-\mathbf{z}^{p}_t $, respectively. We linearize the state and observation dynamics around the nominal trajectories of state and control as $ \mathbf{x}_{t+1}=f(\mathbf{x}_{t},\mathbf{u}_{t},\boldsymbol{\omega}_{t}) \approx \mathbf{x}^{p}_{t+1}+\mathbf{A}_t\tilde{\mathbf{x}}_t + \mathbf{B}_t\tilde{\mathbf{u}}_t +\mathbf{G}_t\boldsymbol{\omega}_t$, and  $ \mathbf{z}_{t}=h(\mathbf{x}_{t},\boldsymbol{\nu}_{t}) \approx \mathbf{z}^{p}_{t}+\mathbf{H}_{t}\tilde{\mathbf{x}}_t+\mathbf{M}_t\boldsymbol{\nu}_t $, where $ \mathbf{A}_t=\partial f(\mathbf{x},\mathbf{u},\boldsymbol{\omega})/\partial \mathbf{x}|_{ \mathbf{x}^{p}_{t}, \mathbf{u}^{p}_{t}, \mathbf{0} } $, $ {\mathbf{B}}_t=\partial f(\mathbf{x},\mathbf{u},\boldsymbol{\omega})/\partial \mathbf{u}|_{ \mathbf{x}^{p}_{t}, \mathbf{u}^{p}_{t}, \mathbf{0} } $, $ {\mathbf{G}}_t=\partial f(\mathbf{x},\mathbf{u}_t,\boldsymbol{\omega})/\partial \boldsymbol{\omega}|_{ \mathbf{x}^{p}_{t}, \mathbf{u}^{p}_{t}, \mathbf{0} } $, $ \mathbf{H}_{t}=\partial h(\mathbf{x},\boldsymbol{\nu})/\partial \mathbf{x}|_{ \mathbf{x}^{p}_{t},\mathbf{0} } $ and $ \mathbf{M}_{t}=\partial h(\mathbf{x},\boldsymbol{\nu})/\partial \boldsymbol{\nu}|_{ \mathbf{x}^{p}_{t},\mathbf{0} } $ are the corresponding Jacobian matrices. Note that, $ \tilde{\mathbf{z}}_t=\mathbf{H}_{t}\tilde{\mathbf{x}}_t+\mathbf{M}_t\boldsymbol{\nu}_t $ denotes the observation innovation, as well.

\textit{Linearization of the belief dynamics:} Corresponding to the nominal trajectory of states, there exists a nominal (Gaussian) trajectory of beliefs, $ \{\mathbf{b}^{p}_t\}_{t=0}^{K} $, where $ \mathbf{x}^{p}_t=\mathbb{E}[\mathbf{b}^{p}_t] $, and the covariance evolution follows the Riccati equations from a KF. Likewise, we linearize the belief dynamics around the nominal trajectories of belief, control and observation as $ \mathbf{b}_{t+1}=\tau(\mathbf{b}_{t},\mathbf{u}_t,\mathbf{z}_{t+1}) \approx \mathbf{b}^{p}_{t+1}+ \mathbf{T}^{\mathbf{b}}_t\tilde{\mathbf{b}}_t+\mathbf{T}^{\mathbf{u}}_t\tilde{\mathbf{u}}_t+ \mathbf{T}^{\mathbf{z}}_t\tilde{\mathbf{z}}_{t+1} $, where $ \mathbf{b}^{p}_{t+1}=\tau(\mathbf{b}^{p}_{t},\mathbf{u}^{p}_t,\mathbf{z}^{p}_{t+1}) $, and $ \mathbf{T}^{\mathbf{b}}_t=\partial \tau(\mathbf{b},\mathbf{u},\mathbf{z})/\partial b|_{\mathbf{b}^{p}_{t}, \mathbf{u}^{p}_{t}, \mathbf{z}^{p}_{t+1} } $, $ \mathbf{T}^{\mathbf{u}}_t=\partial \tau(\mathbf{b},\mathbf{u},\mathbf{z})/\partial \mathbf{u}|_{\mathbf{b}^{p}_{t}, \mathbf{u}^{p}_{t}, \mathbf{z}^{p}_{t+1} } $, and $ \mathbf{T}^{\mathbf{z}}_t=\partial \tau(\mathbf{b},\mathbf{u},\mathbf{z})/\partial \mathbf{z}|_{\mathbf{b}^{p}_{t}, \mathbf{u}^{p}_{t}, \mathbf{z}^{p}_{t+1} } $. Moreover, $ \tilde{\mathbf{b}}_t:=\mathbf{b}_t-\mathbf{b}^{p}_t $ denotes the belief error.

\textit{Linearization of the cost function:} Finally, we linearize the cost function around the nominal trajectories of belief and control actions as $ J\approx J^{p}+\mathbb{E}[\sum_{t=0}^{K-1}(\mathbf{C}^{\mathbf{b}}_t\tilde{\mathbf{b}}_t+ \mathbf{C}^{u}_t\tilde{\mathbf{u}}_t)+ \mathbf{C}^{\mathbf{b}}_K\tilde{\mathbf{b}}_K] $, where $ J^{p}:=\sum_{t=0}^{K-1}c_t(\mathbf{b}^{p}_t,\mathbf{u}^{p}_t)+c_K(\mathbf{b}^{p}_K) $, and we assume continuity of the cost function, and $ \mathbf{C}^{\mathbf{b}}_t=\partial c_t(\mathbf{b},\mathbf{u})/\partial {\mathbf{b}}|_{\mathbf{b}^{p}_{t}, \mathbf{u}^{p}_{t}} $, $ \mathbf{C}^{u}_t=\partial c_t(\mathbf{b},\mathbf{u})/\partial \mathbf{u}|_{\mathbf{b}^{p}_{t}, \mathbf{u}^{p}_{t}} $, and $ \mathbf{C}^{\mathbf{b}}_K=\partial c_K(\mathbf{b})/\partial {\mathbf{b}}|_{\mathbf{b}^{p}_{K}} $. Thus, $ \tilde{J}:= J-J^{p} $ is the error occurred in cost function by our approximation scheme.

\textit{Feedback controller:} As mentioned before, we assume the search is over linear feedback policies, which is a valid assumption for locally controlling a linearized model around a nominal trajectory. Our design, based on the separation principle, supposes the existence of an LQR controller to track and stabilize the trajectory of states around the nominal designed trajectory. Thus, $ \tilde{\mathbf{u}}_{t}=-\mathbf{L}_{t}(\hat{\mathbf{x}}_{t}-\mathbf{x}^{p}_{t}) $. Note, although we are working with the linearized system, the original system is not a linear system and certainty equivalence is only valid with strict limitations on the nonlinearities. Therefore, we cannot replace $ \hat{\mathbf{x}}_{t} $ with $ \mathbf{x}_t $ in the control law without proofs and considerations.

\begin{assumption}[Assumption on calculation of summations]\label{assumption 1}
We assume in the rest of this section for simplicity of our formulas in any sum such as $ \sum_{\tau=t_1}^{t_2}[\mathrm{function}(\tau)] $ that $ t_1\le t_2 $, where $ \mathrm{function} $ is a general function. In other words, $ \mathrm{function}(\tau) $ is only evaluated for $ t_1\le \tau\le t_2 $. Otherwise, it is not calculated, and the summation is zero.
\end{assumption}

\begin{lemma}\textup{\textbf{Estimation Error Propagation}}\label{lemma:Estimation} Define $ \check{\mathbf{x}}_t:=\mathbf{x}_t-\hat{\mathbf{x}}_{t} $ for $ t\ge 0 $ to be the estimation error. Then, for $ t\ge -1 $ the non-recursive estimation error propagation, $ \check{\mathbf{x}}_{t+1} $, in terms of the independent variables, including process and measurement noises and the initial state error $ \tilde{\mathbf{x}}_0=\mathbf{x}_0-\mathbf{x}^{p}_{0} $ can be written as follows:
\begin{align}\label{eq:Estimation Error Propagation}
\check{\mathbf{x}}_{t+1}=\tilde{\mathbf{F}}_{0:t}\tilde{\mathbf{x}}_{0}+\sum\limits_{s=0}^{t}\tilde{\mathbf{F}}_{s+1:t}(\mathbf{U}_{s+1}\mathbf{G}_s\boldsymbol{\omega}_s
-\mathbf{K}_{s+1}\mathbf{M}_{s+1}\boldsymbol{\nu}_{s+1}),
\end{align}
where $ \mathbf{U}_{t+1}:=\mathbf{I}-\mathbf{K}_{t+1}\mathbf{H}_{t+1}, t\ge 0 $, $ \mathbf{F}_{t}:=\mathbf{U}_{t+1}\mathbf{A}_t, t\ge 0 $, and $ \tilde{\mathbf{F}}_{t_1:t_2}:=\Pi_{t=t_1}^{t_2}\mathbf{F}_{t}, t_2\ge t_1\ge 0 $ and otherwise, it is identity matrix.
\end{lemma}

\begin{lemma}\textup{\textbf{State Error Propagation}}\label{lemma:State} Let state error be $ \tilde{\mathbf{x}}_{t}=\mathbf{x}_{t}-\mathbf{x}^{p}_{t} $ for $ t\ge 0 $. Then, for $ t\ge -1 $ the non-recursive state error propagation, $ \tilde{\mathbf{x}}_{t+1} $, in terms of the independent variables, including process and measurement noises and the initial state error $ \tilde{\mathbf{x}}_0 $ can be written as follows:
\begin{align}\label{eq:State Error Propagation}
\tilde{\mathbf{x}}_{t+1}=\tilde{\mathbf{D}}^{\mathbf{x}_{0}}_{t}\tilde{\mathbf{x}}_0+\sum\limits_{s=0}^{t}\tilde{\mathbf{D}}^{\boldsymbol{\omega}}_{s,t}\boldsymbol{\omega}_s
-\sum\limits_{s=0}^{t-1}\tilde{\mathbf{D}}^{\boldsymbol{\nu}}_{s+1,t}\boldsymbol{\nu}_{s+1},
\end{align}
where $ \tilde{\mathbf{F}}^{\mathbf{x}_0}_{t+1}:=\mathbf{L}_{t+1}\tilde{\mathbf{F}}_{0:t}, t\ge -1 $, $ \tilde{\mathbf{F}}^{\boldsymbol{\omega}}_{s,t+1}:=\mathbf{L}_{t+1}\tilde{\mathbf{F}}_{s+1:t}\mathbf{U}_{s+1}\mathbf{G}_s, t\ge 0, t\ge s\ge 0 $, $ \tilde{\mathbf{F}}^{\boldsymbol{\nu}}_{s+1,t+1}:=\mathbf{L}_{t+1}\tilde{\mathbf{F}}_{s+1:t}\mathbf{K}_{s+1}\mathbf{M}_{s+1},t\ge 0, t\ge s\ge 0 $, $ \mathbf{D}_{t}:=\mathbf{A}_t- \mathbf{B}_t\mathbf{L}_t, t\ge 1 $, $ \mathbf{D}_{0}:=\mathbf{A}_0 $ and $ \tilde{\mathbf{D}}_{t_1:t_2} = \Pi_{t=t_1}^{t_2}\mathbf{D}_{t}, t_2\ge t_1\ge 0 $ otherwise, it is identity matrix. Moreover, $ \tilde{\mathbf{F}}^{\boldsymbol{\omega}}_{s,r,t}:=\tilde{\mathbf{D}}_{r+1:t}\mathbf{B}_r\tilde{\mathbf{F}}^{\boldsymbol{\omega}}_{s,r}, t-1\ge s\ge 0, t\ge r\ge s+1, t\ge 1 $, $ \tilde{\mathbf{F}}^{\boldsymbol{\nu}}_{s+1,r,t}:=\tilde{\mathbf{D}}_{r+1:t}\mathbf{B}_r\tilde{\mathbf{F}}^{\boldsymbol{\nu}}_{s+1,r}, t-1\ge s\ge 0, t\ge r\ge s+1, t\ge 1 $, $ \tilde{\mathbf{D}}^{\mathbf{x}_{0}}_{t}:=\tilde{\mathbf{D}}_{0:t}+\sum\limits_{r=1}^{t}\tilde{\mathbf{D}}_{r+1:t}\mathbf{B}_r\tilde{\mathbf{F}}^{\mathbf{x}_0}_{r}, t\ge -1 $, $ \tilde{\mathbf{D}}^{\boldsymbol{\omega}}_{s,t}:=\tilde{\mathbf{D}}_{s+1:t}\mathbf{G}_s+\sum\limits_{r=s+1}^{t}\tilde{\mathbf{F}}^{\boldsymbol{\omega}}_{s,r,t}, 1\le s\le t-1, t\ge 1 $, $ \tilde{\mathbf{D}}^{\boldsymbol{\omega}}_{t,t}:=\tilde{\mathbf{D}}_{t+1:t}\mathbf{G}_t=\mathbf{G}_t, t\ge 0 $, $ \tilde{\mathbf{D}}^{\boldsymbol{\omega}}_{0,t}:=\sum\limits_{r=1}^{t}\tilde{\mathbf{F}}^{\boldsymbol{\omega}}_{0,r,t}, t\ge 1 $, and $ \tilde{\mathbf{D}}^{\boldsymbol{\nu}}_{s+1,t}:=\sum\limits_{r=s+1}^{t}\tilde{\mathbf{F}}^{\boldsymbol{\nu}}_{s+1,r,t}, 0\le s\le t-1, t\ge 1 $.
\end{lemma}

\begin{lemma}\textup{\textbf{Control Error Propagation}}\label{lemma:Control} Let control error be $ \tilde{\mathbf{u}}_{t}=\mathbf{u}_{t}-\mathbf{u}^{p}_{t} $ for $ t\ge 0 $. Then, for $ t\ge -1 $ the non-recursive control error propagation, $ \tilde{\mathbf{u}}_{t+1} $, in terms of the independent variables, including process and measurement noises and the initial state error $ \tilde{\mathbf{x}}_0 $ can be written as follows:
\begin{align*}
\tilde{\mathbf{u}}_{t+1}=-\mathbf{L}^{\mathbf{x}_{0}}_{t+1}\tilde{\mathbf{x}}_0
-\sum\limits_{s=0}^{t}\mathbf{L}^{\boldsymbol{\omega}}_{s,t+1}\boldsymbol{\omega}_s
-\sum\limits_{s=0}^{t}\mathbf{L}^{\boldsymbol{\nu}}_{s,t+1}\boldsymbol{\nu}_{s+1},
\end{align*}
where $ \mathbf{L}^{\mathbf{x}_{0}}_{t+1}:=\mathbf{L}_{t+1}\tilde{\mathbf{D}}^{\mathbf{x}_{0}}_{t}-\tilde{\mathbf{F}}^{\mathbf{x}_0}_{t+1}, t\ge 1 $, $ \mathbf{L}^{\boldsymbol{\omega}}_{s,t+1}:=\mathbf{L}_{t+1}\tilde{\mathbf{D}}^{\boldsymbol{\omega}}_{s,t}-\tilde{\mathbf{F}}^{\boldsymbol{\omega}}_{s,t+1}, t\ge 1, t\ge s\ge 0 $, $ \mathbf{L}^{\boldsymbol{\nu}}_{s,t+1}:=\tilde{\mathbf{F}}^{\boldsymbol{\nu}}_{s+1,t+1}-\mathbf{L}_{t+1}\tilde{\mathbf{D}}^{\boldsymbol{\nu}}_{s+1,t}, t\ge 1, t-1\ge s\ge 0 $ and $ \mathbf{L}^{\boldsymbol{\nu}}_{t,t+1}:=\tilde{\mathbf{F}}^{\boldsymbol{\nu}}_{t+1,t+1} $.
\end{lemma}

\begin{lemma}\textup{\textbf{Observation Error Propagation}}\label{lemma:Observation} Let observation error be $ \tilde{\mathbf{z}}_{t}=\mathbf{z}_{t}-\mathbf{z}^{p}_{t} $ for $ t\ge 1 $. Then, for $ t\ge 0 $ the non-recursive observation error propagation, $ \tilde{\mathbf{z}}_{t+1} $, in terms of the independent variables, including process and measurement noises and the initial state error $ \tilde{\mathbf{x}}_0 $ can be written as follows:
\begin{align}
\tilde{\mathbf{z}}_{t+1}=\tilde{\mathbf{H}}^{\mathbf{x}_{0}}_{t+1}\tilde{\mathbf{x}}_0+\sum\limits_{s=0}^{t}\tilde{\mathbf{H}}^{\boldsymbol{\omega}}_{s,t+1}\boldsymbol{\omega}_s
+\sum\limits_{s=0}^{t}\tilde{\mathbf{H}}^{\boldsymbol{\nu}}_{s+1,t+1}\boldsymbol{\nu}_{s+1},
\end{align}
where $ \tilde{\mathbf{H}}^{\mathbf{x}_{0}}_{t+1}:=\mathbf{H}_{t+1}\tilde{\mathbf{D}}^{\mathbf{x}_{0}}_{t}, t\ge 0 $, $ \tilde{\mathbf{H}}^{\boldsymbol{\omega}}_{s,t+1}:=\mathbf{H}_{t+1}\tilde{\mathbf{D}}^{\boldsymbol{\omega}}_{s,t}, t\ge 0, t\ge s\ge 0 $, $ \tilde{\mathbf{H}}^{\boldsymbol{\nu}}_{s+1,t+1}:=-\mathbf{H}_{t+1}\tilde{\mathbf{D}}^{\boldsymbol{\nu}}_{s+1,t}, t\ge 1, t-1\ge s\ge 0 $, and $ \tilde{\mathbf{H}}^{\boldsymbol{\nu}}_{t+1,t+1}:=\mathbf{M}_{t+1}, t\ge 0 $.
\end{lemma}

\begin{lemma}\textup{\textbf{Belief Error Propagation}}\label{lemma:Belief} Let belief error be $ \tilde{\mathbf{b}}_{t}=\mathbf{b}_{t}-\mathbf{b}^{p}_{t} $ for $ t\ge 0 $. Then, for $ t\ge 0 $ the non-recursive belief error propagation in terms of the independent variables, including process and measurement noises and the initial state error $ \tilde{\mathbf{x}}_0 $ can be written as follows:
\begin{align}
\tilde{\mathbf{b}}_t=\tilde{\mathbf{T}}^{\mathbf{x}_{0}}_{t}\tilde{\mathbf{x}}_0
+\sum\limits_{s=0}^{t-1}\tilde{\mathbf{T}}^{\boldsymbol{\omega}}_{s,t}\boldsymbol{\omega}_s
+\sum\limits_{s=1}^{t}\tilde{\mathbf{T}}^{\boldsymbol{\nu}}_{s,t}\boldsymbol{\nu}_{s},
\end{align}
where $ \tilde{\mathbf{T}}^{\mathbf{b}}_{t_1:t_2}:=\Pi_{t=t_1}^{t_2}\tilde{\mathbf{T}}^{\mathbf{b}}_{t}, 0\le t_1\le t_2 $ otherwise, it is identity matrix, $ \tilde{\mathbf{T}}^{\mathbf{x}_{0}}_{t}:=\mathbf{T}^{\mathbf{z}}_{t-1}\tilde{\mathbf{H}}^{\mathbf{x}_{0}}_{t}+\sum\limits_{s=0}^{t-2}(-\tilde{\mathbf{T}}^{\mathbf{b}}_{s+2:t-1}\mathbf{T}^{\mathbf{u}}_{s+1}\mathbf{L}^{\mathbf{x}_{0}}_{s+1}+\tilde{\mathbf{T}}^{\mathbf{b}}_{s+1:t-1}\mathbf{T}^{\mathbf{z}}_s\tilde{\mathbf{H}}^{\mathbf{x}_{0}}_{s+1}), t\ge 1 $, $ \tilde{\mathbf{T}}^{\mathbf{x}_{0}}_{0}:=\mathbf{0} $, $ \tilde{\mathbf{T}}^{\boldsymbol{\omega}}_{s,t}:=\mathbf{T}^{\mathbf{z}}_{t-1}\tilde{\mathbf{H}}^{\boldsymbol{\omega}}_{s,t}+\sum\limits_{r=s}^{t-2}(-\tilde{\mathbf{T}}^{\mathbf{b}}_{r+2:t-1}\mathbf{T}^{\mathbf{u}}_{r+1}\mathbf{L}^{\boldsymbol{\omega}}_{s,r+1}+\tilde{\mathbf{T}}^{\mathbf{b}}_{r+1:t-1}\mathbf{T}^{\mathbf{z}}_r\tilde{\mathbf{H}}^{\boldsymbol{\omega}}_{s,r+1}), t\ge 2, t-2\ge s\ge 0 $, $ \tilde{\mathbf{T}}^{\boldsymbol{\omega}}_{t-1,t}:=\mathbf{T}^{\mathbf{z}}_{t-1}\tilde{\mathbf{H}}^{\boldsymbol{\omega}}_{t-1,t}, t\ge 1 $, $ \tilde{\mathbf{T}}^{\boldsymbol{\nu}}_{s,t}:=\mathbf{T}^{\mathbf{z}}_{t-1}\tilde{\mathbf{H}}^{\boldsymbol{\nu}}_{s,t}+\sum\limits_{r=s-1}^{t-2}(-\tilde{\mathbf{T}}^{\mathbf{b}}_{r+2:t-1}\mathbf{T}^{\mathbf{u}}_{r+1}\mathbf{L}^{\boldsymbol{\nu}}_{s-1,r+1}+\tilde{\mathbf{T}}^{\mathbf{b}}_{r+1:t-1}\mathbf{T}^{\mathbf{z}}_r\tilde{\mathbf{H}}^{\boldsymbol{\nu}}_{s,r+1}), t\ge 2, t-1\ge s\ge 1 $, and $ \tilde{\mathbf{T}}^{\boldsymbol{\nu}}_{t,t}:=\mathbf{T}^{\mathbf{z}}_{t-1}\tilde{\mathbf{H}}^{\boldsymbol{\nu}}_{t,t}, t\ge 1 $.
\end{lemma}

\begin{theorem}\textup{\textbf{Cost Function Error}}\label{theroem:Cost} Let cost function error be $ \tilde{J}=J-J^{p} $ for $ t\ge 0 $. Given that process and observation noises are zero mean i.i.d. and are mutually independent from each other and the initial belief, under a first-order approximation, the stochastic cost function is dominated by the nominal part of the cost function. Moreover the expected first-order error is zero, i.e., $ \tilde{J}=0 $.
\end{theorem}

\textit{Discussion:} Theorem \ref{assumption 1} shows that under the assumption that the error resulting from linearizations are small enough (i.e., the linearizations are valid), the error in our cost function is independent from the stochastic terms, and it is inconsequential. Therefore, under these conditions, the original cost function is dominated by the nominal part of the cost function. In practice, the time horizon is only chosen large enough to find feasible solutions, and the linearization error is negligible. Thus, choosing the underlying linearization trajectory (or equivalently, the control actions corresponding to the nominal trajectory) as the optimization variables, the optimal underlying trajectory can be calculated. Moreover, whenever the accumulated error of the belief approximation under these assumption gets higher than a threshold, the problem restarts and replanning follows. It is important to note that the separation principle is the central idea behind the method, since it provides the mechanism to design the controller and estimator separate from each other. The T-LQG method, utilizes this theory and the dependence of the LQG on the underlying trajectory, and through a coupled design of trajectory with the provided LQG methodology and a standard optimization problem, finds the LQG controller with the best performance. 

Hence, utilizing the separation theorem makes all these arguments possible, because, there is a feedback policy that changes the sensitivity matrices of the error propagations and stabilizes the system. Otherwise, the control error grows intractably and becomes an independent uncontrollable variable, thus, there is no guarantee for our statements. Moreover, since the LQR feedback gain is only dependent on the underlying trajectory (and is independent from the estimation by separation), the trajectory optimization is performed without the feedback, and the controller is designed on top of the policy.

\section{Comparison of Methods}\label{sec:Comparison of Methods}
In this section, we provide a comparison between state-of-the-art belief space planning approaches from a methodology and computational complexity perspective. We make occasional references to the following methods: a) LQG-MP \cite{van2011lqg}, b) iLQG-based method \cite{van2012motion}, c) SELQR \cite{sun2016stochastic} d) the method utilizing MLO \cite{Platt10}, e) the non-Gaussian Receding Horizon Control (RHC)-based method \cite{platt2013convex}, f) the non-Gaussian observation covariance reduction method \cite{rafieisakhaei2016feedbackICRA}, g) FIRM \cite{Ali14}, and g) the point-based POMDP solvers \cite{Sondik71,Pineau03,GShaniJPineau13,seiler2015online}. Table \ref{table:comparison} summarizes the key differences between the methods.

\begin{sidewaystable}[!]\centering
\ra{1.3}\caption{Comparison of belief space planning methods on important issues.} \label{table:comparison}
\begin{tabular}{@{}c c c c c c@{}}\toprule
	&	\multicolumn{1}{m{2.5cm}}{\centering Planning as an Optimization}	&	\multicolumn{1}{m{3.5cm}}{\centering Linearization Trajectory (Exploitable for Optimization)}	&	Planning Observations	&	\multicolumn{1}{m{3.3cm}}{\centering Computational Complexity}	&	Convergence	Rate
	
\\\midrule
LQG-MP \cite{van2011lqg}	&	None	&	\multicolumn{1}{m{3.5cm}}{\centering RRT trajectories (No)}	&	---	&	$O(n_rKn^3)$	&	---	

\\ iLQG-based \cite{van2012motion}	&	DP	&	\multicolumn{1}{m{3.5cm}}{\centering Fixed at each iteration (No)}	&	Stochastic observations	&	$O(Kn^6)$	&	\multicolumn{1}{m{2.8cm}}{\centering Second order (line-search tuning)}	

\\SELQR \cite{sun2016stochastic}	&	DP	&	\multicolumn{1}{m{3.5cm}}{\centering Fixed at each iteration (No)}	&	MLO	&	$O(Kn^6)$	&	Second order	

\\\multicolumn{1}{m{2.8cm}}{\centering MLO \cite{Platt10}}	&	NLP	&	\multicolumn{1}{m{3.5cm}}{\centering Predicted mean update (Yes)}	&	MLO	&	\multicolumn{1}{m{3.3cm}}{\centering $O(n_{tr}(Kn^3\!+\!kn^2))$ or $O(n_{tr}(Kn^3\!+kn^2)\!+\! Kn^6)$}	&	SQP rate	

\\\multicolumn{1}{m{2.8cm}}{\centering Non-Gaussian RHC-Based \cite{platt2013convex}}	&	Convex	&	\multicolumn{1}{m{3.5cm}}{\centering Linear propagation of initial estimate (Yes)}	&	\multicolumn{1}{m{3.5cm}}{\centering MLO}	&	$O(NK(Kn^3\!+\!Nn^2))$	&	$\Omega((N\!+\!Kn)\log(\frac{1}{\epsilon}))$	

\\\multicolumn{1}{m{2.8cm}}{\centering Non-Gaussian Obs. Cov. Reduction \cite{rafieisakhaei2016feedbackICRA}}	&	Convex	&	\multicolumn{1}{m{3.5cm}}{\centering Linear propagation of initial estimate (Yes)}	&	\multicolumn{1}{m{3.5cm}}{\centering Predicted ensemble of observation particles}	&	$O(Kn^3\!+\!Nn^2)$	&	$\Omega(Kn\log(\frac{1}{\epsilon}))$	

\\T-LQG	&	NLP	&	\multicolumn{1}{m{3.5cm}}{\centering Non-linear propagation of initial estimate (Yes)}	&	---	&	$O(Kn^3)$	&	Second order	\\
\bottomrule 
\end{tabular}\par\noindent
\begin{itemize}\compresslist
\item[\textbullet] We assume the size of vectors $ \mathbf{x}, \mathbf{u} $ and $ \mathbf{z} $ are all $ O(n) $, and $ K $ is planning horizon
\item[\textbullet] $ n_r $ is the number of RRT paths generated in \cite{van2011lqg}
\item[\textbullet] For the method of \cite{Platt10}, $ n_{tr} $ is the number of transcription steps in the direct transcription; $ k $ is the number of unit vectors pointing in the desired directions to minimize the covariance in; the second computational complexity is valid if the B-LQR is also used, otherwise, the first complexity is more accurate
\item[\textbullet] $ N $ is the number of samples, $ \epsilon $ is the convergence error \item[\textbullet] \textit{Convergence Rate} is the number of calls needed to the oracle to converge using a method such as center of gravity 
\item[\textbullet] \textit{DP} is Dynamic Programming
\item[\textbullet] \textit{Second order rate} is the general rate for Newton-like methods
\end{itemize}
\end{sidewaystable}

As reflected in the table, a central difference between these methods is the way the system and observation equations are linearized. After linearization of the equations, the corresponding Jacobians become coupled with the trajectory. Therefore, if the underlying linearization trajectory is a sequence of fixed points, the Jacobians become constant matrices for the entire optimization, and the structure of the system models (on which depends many other properties of the system, such as sensitivity of the observations, controllability, reachability, etc.) essentially become fixed, untouchable, and, more importantly, un-exploitable for the optimization purposes. Table \ref{table:comparison} summarizes the capability of methods on using this feature. As noted, our method fully exploits this property and finds the best linearization trajectory among the methods. Moreover, assumptions on observations in our method are inconsequential and the observation model is exploited with its best capacity. Most importantly, the computational complexity of T-LQG is the lowest among all. 

Note: the computational complexity only reflects the calculations of the core problems for belief space planning in each method. For obstacle-avoidance, each method has a different approach, which is out of this discussion and can be further detailed in a pure motion-planning scope. The information in table \ref{table:comparison} and the calculations regarding the computational complexity are estimated to the best of our knowledge. 

Next, we provide a brief summary of the methods and afterwards, we elaborate more on the key methodological aspects and differences.
\subsection{An Overall Summary of the Methods}
\textbf{a) LQG-MP \cite{Berg11-IJRR}} In this method, several paths generated by RRT planner are taken as initial nominal trajectories, and the system equations are linearized around those trajectories. An LQG tracker is designed along each trajectory and the control sequences are compared based on an obstacle-avoidance performance measure. The trajectory with the best performance is selected as the nominal trajectory to track and the LQG tracker corresponding to that trajectory is chosen as the policy to implement.

\textbf{b) iLQG-based method \cite{van2012motion}} In this method, the iteration begins with an initial guess trajectory that is obtained using a method such as RRT, around which the system equations, belief dynamics and value function are linearized. Then, the value function is evaluated by backward run along the nominal trajectory. Next, the noiseless belief dynamics is used to forward propagate the belief using the policy that was found in the backward propagation. This gives a new nominal trajectory for the next iteration of the algorithm. The iterations are coupled with an adaptive line search method and continue until convergence to a locally optimal policy.

\textbf{c) SELQR \cite{sun2016stochastic}} In these methods, the iteration idea of the iLQG-based methods is extended by a better choice of the underlying linearization trajectory. Starting with an initial guess, the forward and backward iterations are both done over that trajectory, then sum of the costs of forward and backward iterations at every time step is obtained. This defines a minimization problem whose result provides the nominal trajectory for linearization in the next iteration.

\textbf{d) MLO \cite{Platt10}} This method is also based on the LQG methodology. The mean update equation in the (extended) Kalman filtering equation requires an observation (or an assumption over the observations) to calculate the innovation term, whereas the covariance update equation only depends on an underlying trajectory (this trajectory can either come from the true mean update during the estimation, or can be a fixed nominal trajectory). Moreover, the mean update equations are tied to the covariance update, as well. In this method, in order to perform the mean update, the future observations are assumed to be the most-likely observations (which correspond to the noiseless observations predicted by the observation model). The system equations are linearized around such mean updates at each step. An optimization problem with a quadratic cost is defined to obtain the desired trajectory, and an LQR controller is used to reject the disturbances.

\textbf{e) Non-Gaussian RHC-Based \cite{platt2013convex}} In this method, the most-likely observation method is adapted for a linear system and observation models with Gaussian noises, where the observation noise covariance is state-dependent. The representation of the belief is replaced with that of a particle filter, and the noise models are utilized to obtain the dynamics of the particle weights. An optimization problem is defined and convexified to obtain the optimal nominal trajectory. The policy is implemented with an RHC strategy closing the feedback loop in the execution.

\textbf{f) Non-Gaussian Observation Covariance Reduction \cite{rafieisakhaei2016feedbackICRA}} In this method, system equations are linearized around an initial nominal trajectory, however, the observation model is linearized around the noiseless propagation of the initial estimate. The main contribution of this work is to exploit the observation uncertainty and define an optimization problem which is easy to solve, avoids performing the filtering equations and yields similar trajectories as the other belief space planning methods. Moreover, the belief has a particle filter representation where no assumptions on the noise distributions are assumed.

\textbf{g) FIRM \cite{Ali14}} Feedback Information RoadMap is an offline POMDP planner that solves an MDP over a graph with finite number of nodes in the belief space. Therefore, the solution over the graph is provided based on the dynamic programming. As mentioned before, in the point-based POMDP solvers where the probability of reaching to a belief node is zero and whence the solution is only valid for the initial belief. Unlike the point-based solvers, the key point in FIRM is stabilizing the belief over a belief node in the graph with high probability utilizing an stabilizer controller. This, also breaks the curse of history. Currently the abstracted algorithm of FIRM has been implemented utilizing the LQG methodology and is called the SLQG-FIRM.

\textbf{h) Point-Based POMDP Solvers \cite{Sondik71,Pineau03,GShaniJPineau13,seiler2015online}} The POMDP problem was introduced in 1971 in \cite{Sondik71}, with an algorithm to obtain the exact optimal solution using the alpha-vectors. The algorithm then evolved into an anytime algorithm in 2003 in \cite{Pineau03}, introducing the point-based POMDP solvers. This method has been the foundation for the majority of research in the POMDP field \cite{GShaniJPineau13}. There has been many successes in finding solving POMDP benchmark problems with low CPU-times. Even the latest advancements in the field, such as \cite{seiler2015online}, suffer from multiple limitations. For instance, the scalability with time-horizon seems to be a fundamental limitation that might be difficult to just overcome. Ad-hoc solutions to reduce the planning time horizon to local planning (which are much lower than enough for reaching the goal region) and replanning every few steps is not a feasible a solution for practical problems. This is also much different from the Model Predictive Control (MPC) strategy, where the planning horizon is chosen to be large enough to reach to a goal region.

An issue of POMDP solvers is that the search over the belief space is reduced to a discrete set of belief nodes (either through a discretization of the underlying spaces or through random sampling of continuous spaces and building a decision tree over belief samples). In these methods, the probability of re-visiting any particular discrete belief node in the tree (other than the root) is equal to zero. Thus, the solution is \textit{only} valid for the initial belief. A way of overcoming this limitation is to perform a continuous branching or an exact Monte-Carlo, where for every infinitesimal change in a higher level of the tree, there is an exponentially increased number of belief nodes in the next level, which brings back the original highly computational theoretical solution of POMDPs. It is only in such a case where the solution is comparable to methods such as ours, where the search occurs over a continuous set of beliefs---thus, the replanning does not need to happen every single step. For this reason, our solution is valid for a much longer horizon and for a belief space region far more considerable than results from point-based POMDP solvers. 

Moreover, in T-LQG, by tracking the nominal and true belief during online implementation, whenever the optimality deviation is more than the tolerable threshold, replanning occurs, which is essentially impractical in true long-horizon POMDP solvers. FIRM \cite{Ali14} on the other hand, provides an offline approach to tackle the original POMDP problem by solving the dynamic programming over a graph in the belief space and breaking the curse of history; but, to get closer to optimality, more FIRM nodes need to be sampled in the offline design.

\subsection{Comparison on Important Issues}
In this section we discuss more on the key differences between methods (a-f). Since, POMDPs were already discussed before, we avoid further discussions in here. Moreover, since FIRM is an offline planner, we do not compare with FIRM either. We explain how the linearization trajectory is different in these methods and how that leads to major differences in the algorithms. Moreover, we explain that a critical difference is the assumptions on the observation process during the planning stage. Note that, likewise methods (a-d), our current paper deals with Gaussian beliefs.

\textit{Optimization problem:} In (a), the least-cost trajectory is chosen among a finitely generated initial trajectories, hence the underlying trajectory is not optimized or morphed. In (b), the underlying trajectory is morphed through an iteration mentioned as above coupled with tuning of a line-search method. Thus, the algorithm does not involve an explicit optimization problem that can be solved via an NLP solver. Rather, the whole method involves the inner mechanisms of an optimization problem. The method is essentially a dynamic-programming-based algorithm. Therefore, the merits of an explicit NLP problem cannot be exploited. In (c), the approach is similar to (b), with a difference that there is also an intermediary optimization problem in each back and forth iteration to find a better nominal trajectory for the next iteration. However, the whole algorithm is essentially similar in content to the method of (b) and the problem lacks a standard optimization problem. In (d) the trajectory optimization problem is posed as an optimization problem that can be solved using SQP. In (e) and (f), the problem is convexified and can be solved using any convex optimizer. Our method also presents the planning problem as an NLP program that can be solved by a generic NLP solver. Presenting the problem as an standard optimization problem has the advantage that it can be solved using various tools and softwares in the optimization and control theory, increasing the efficiency of implementation and availing the usage of advanced techniques developed in those fields to obtain smoother solutions. Moreover, it does not require delving into the details of optimization problem solving.

\textit{Linearization of the system equations:} As pointed above, this is a central difference between the methods. Essentially, an LQG planner with a form of Kalman filtering for estimation requires a nominal trajectory to linearize the system equations. As mentioned before, after linearization of the equations, the Jacobians correspond to the specific trajectory. Therefore, if the underlying linearization trajectory is not a variable of optimization, the Jacobians become constant matrices for the entire optimization and un-exploitable for the optimization purposes. This is what happens in methods (a), (b), and (c). In these methods, although, the underlying linearization changes during the whole algorithm; however, the linearization of the equations is decoupled from the manipulations and deformations of the underlying trajectory, and they happen sequentially with respect to each other. In (e), the model is linear to begin with. On the other hand, in (d) and (f), the linearization is coupled with the manipulation of the trajectory. However, methods are different; in (e), the linearization is done over the predicted mean of the belief (whose updates are possible based on most-likely observations assumption), but in (f), the underlying trajectory for the observation model is the parametrized possible trajectories obtained from the noiseless propagation of the initial estimate, and the trajectory for system equations is based on an initial guess. In this paper, the underlying linearization trajectory is the optimization variable.

\textit{Assumptions on the observation during planning:} The observation distributions are calculated in the methods (a) and (b) based on the LQG methodology; however, in (a), the observations do not contribute to the designed trajectory. In (b), the stochasticity of the observations (distributed with a Gaussian density) is exploited in the dynamic programming equations. In (c), (d) and (e), the observations are most-likely observations. In (f), an ensemble of observation particles for the entire path is generated and their predicted covariance is reduced as an objective in the optimization problem. In the current work, any assumption on the observations is inconsequential and the planning is performed only utilizing the trajectory-dependent Jacobian of the observation model.

\textit{Optimization problem time-complexity for obstacle-free case:} As mentioned, we provide the time complexity for methods (a), (c), (d) and (e) to the best of our knowledge. Let us assume for simplicity that the size of $ \mathbf{x} $,  $\mathbf{u} $, and $ \mathbf{z} $ vectors are all $ O(n) $. The computation time in method a is on finding as many RRT plans as possible, therefore, since this method is not constructing a path the quality of solution can be significantly poorer than the other methods. If $ n_r $ number of RRT paths are taken, then it would take $ O(n_rKn^3) $, however, there is no issue of convergence in here. In (b) and (c), the computation complexity is $ O(K n^6) $ with a second-order convergence rate of Newton-like methods to a locally optimal solution. However, method (c), converges faster than (b), as stated in (c). Method (d), takes $ O(n_{tr}(Kn^3+kn^2)) $, where $ n_{tr} $ is the number of transcription steps in the direct transcription, and $ k $ is the number of unit vectors pointing in $ k $ directions to minimize the covariance in their algorithm. In method (e), utilizing a common method, such as center of gravity for convex optimization \cite{bubeck2014theory} to obtain a \textit{globally optimal} solution with $ \epsilon $ confidence and $ N $ number of samples, the algorithm requires $ O(NK(Kn^3+Nn^2)) $ computations and the convergence needs $ \Omega((N+Kn)log(1/\epsilon)) $ calls to the oracle. In method (f), the convex problem requires $ O(Kn^3+Nn^2) $ computations and $ \Omega(Knlog(1/\epsilon)) $ calls to the oracle \cite{nemirovsky1983problem}. Our current method requires $ O(Kn^3) $ computations and the convergence rate is the rate for the particular gradient-descent method utilized. For instance, a Newton-like method converges at a second-order rate.

\subsection{Comparison on Other Issues}
In this section, we point out some other differences between the methods that are of less importance than the previous points.

\textit{Parametrization of the belief:} In a Gaussian model, it is assumed belief is fully parametrized by two parameters. In a non-Gaussian method this assumption is lifted and typically replaced by a number of samples taken from the belief. Methods (a-d) assume Gaussian beliefs and methods (e) and (f) assume a non-Gaussian representation of the belief. In (e), the particle weights become part of the optimization variables, whereas in (f), the samples or their weights are not variables and the optimization shows more scalability. The Gaussianity assumption can be a valid assumption in the vicinity of a nominal trajectory. Our current paper, deals with Gaussian beliefs. The Gaussianity assumption can be a valid assumption in the vicinity of a nominal trajectory. Therefore, a method that can better stabilize around a nominal trajectory can better exploit this feature. In particular, our method with a better promised path and coupled with feedback controller fully exploits this feature, making the Gaussianity assumption more valid.

\textit{Form of the system equations:} In all methods except (e), the system and observation models are non-linear. In (e), both equations are linear. Moreover, in (d), the process noise is not included.


\textit{Replanning policy:} In (a), (b), and (c), replanning is not discussed. In (d) it is based on the mean deviation from a predicted mean. In (e), a combination of KL divergence and RHC strategy is assumed, and in (f), ar every stage replanning is performed. In our current method, a symmetric distance based on KL divergence is utilized.

\textit{Initialization of the optimization problem:} The initial guess in (a), (b) and (c) is based on an RRT or a similar planner. However, in (a), essentially there is no construction of the path, whereas in the other methods, a path is constructed. In (b), it requires an adaptive line-search and a feasible initial path to ensure convergence. In (d), the optimization yields a locally optimal solution. In (e) and (f) the convex planning problems require no initialization and the planning results are global in the sense of the defined optimization problem. In the current paper, the non-linear optimization requires initialization based on an RRT or a similar planner, and the result of the optimization is a locally optimal path.

\textit{Non-convex constraints:} In (a), a performance measure based on obstacle avoidance is defined to compare the safety of the resulting policies. In (b), (c) and (f), a cost function is added to the optimization problem. In (d), obstacles are not considered. In (e), mixed integer programming and chance constraints are used to avoid constraints. In terms of the computation complexity, among methods (b-f), the methods (b), (c) and (f) have lower computation complexities. In the current paper, an extended version of the method in (f) is introduced, which provides safety based on the barrier functions.

\begin{figure}[t!]
\centering
   \includegraphics[width=0.75\linewidth]{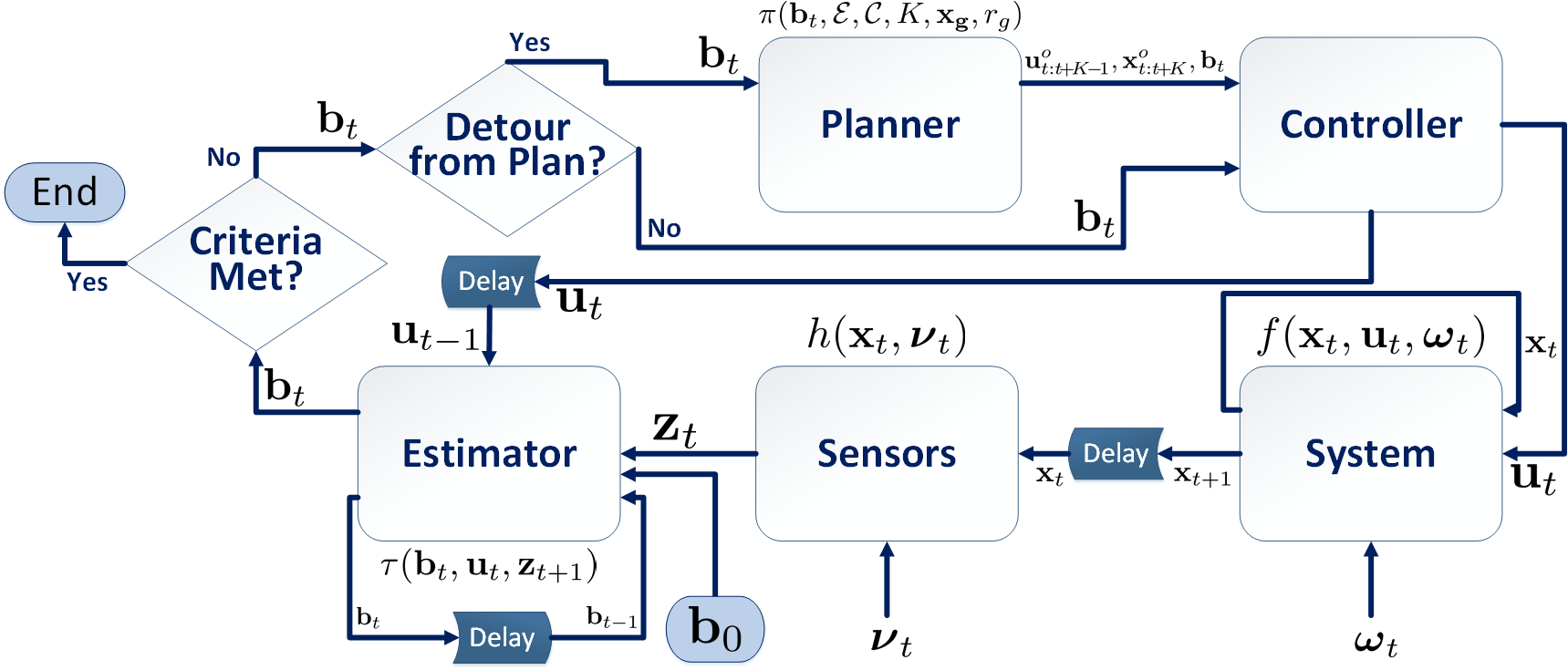}

  \caption{The overall feedback control loop.}
  \label{fig:control_Loop} 
\end{figure}
\begin{algorithm}[b!]
    \SetKwInOut{Input}{Input}
    \Input{Initial belief $\mathbf{b}_{0}$, Goal region $\mathbf{B}_{r_g}(\mathbf{x}_{g})$, Planning horizon $K$, Obstacle parameters $ (\mathcal{E},\mathcal{C}) $}
    $ t\gets0 $\;
    \While{$\mathcal{P}(\mathbf{b}_t,r_g,\mathbf{x}_g)\le p_g$}{
    \If{$ d(\mathbf{b}_t, \mathbf{b}^{o}_t)>d_{th}~ \mathrm{\mathbf{or}}~ t==0~ \mathrm{\mathbf{or}}~ t==K $ }{
    Optimal Trajectory: $ \{\mathbf{u}^o_{0:K\!-\!1},\mathbf{x}^o_{0:K}\}\gets\pi(\mathbf{b}_0, \mathcal{E},\mathcal{C}, K, \mathbf{x_g}, r_g) $\;
        $ t\gets0 $\;
    }\Else{
    Policy Function: $ \hat{\mathbf{x}}_{t}\gets\mathbb{E}[\mathbf{b}_t] $, 
    $ \mathbf{u}_{t}\gets-\mathbf{L}_{t}(\hat{\mathbf{x}}_{t}-\mathbf{x}^{o}_{t}) + \mathbf{u}^{o}_{t} $\;
	Execution: $ \mathbf{x}_{t+1}\gets f(\mathbf{x}_{t},\mathbf{u}_{t},\boldsymbol{\omega}_{t}) $\;
	Perception: $ \mathbf{z}_{t+1}\gets h(\mathbf{x}_{t+1},\boldsymbol{\nu}_{t+1}) $\;
    Estimation: $\mathbf{b}_{t+1}\gets\tau(\mathbf{b}_{t},\mathbf{u}_{t},\mathbf{z}_{t+1})$\;
    $ t\gets t+1 $\;}
    }
\caption{T-LQG}\label{alg:re-planning}
\end{algorithm}

\begin{figure}[t!]
\centering
   \subfloat[{Range and bearing}]{\includegraphics[width=0.32\linewidth]{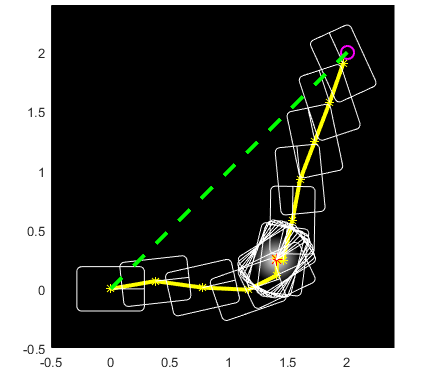}}
   \subfloat[Bearing-only]{\includegraphics[width=0.32\linewidth]{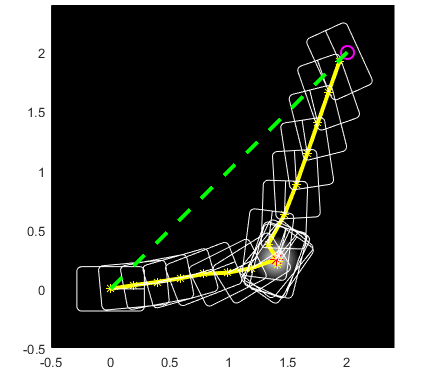}}
   \subfloat[Range-only]{\includegraphics[width=0.32\linewidth]{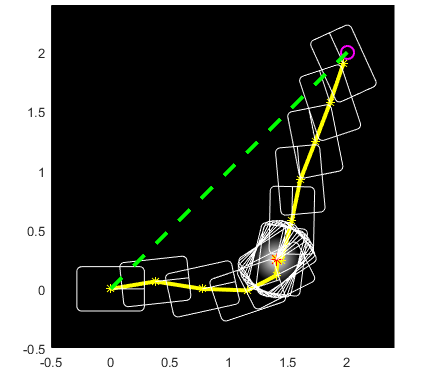}}\\
   \subfloat[Range and bearing]{\includegraphics[width=0.32\linewidth]{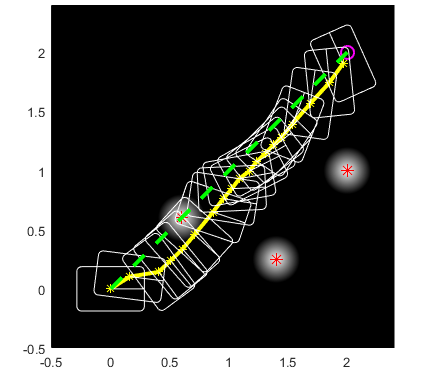}}
   \subfloat[Bearing-only]{\includegraphics[width=0.32\linewidth]{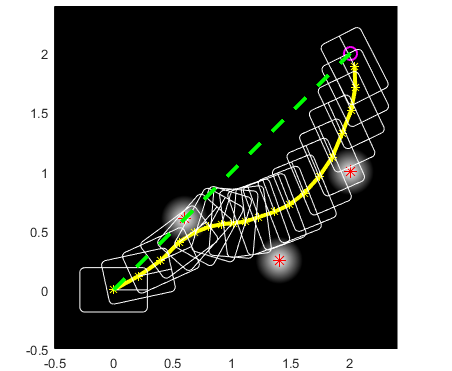}}
   \subfloat[Range-only]{\includegraphics[width=0.32\linewidth]{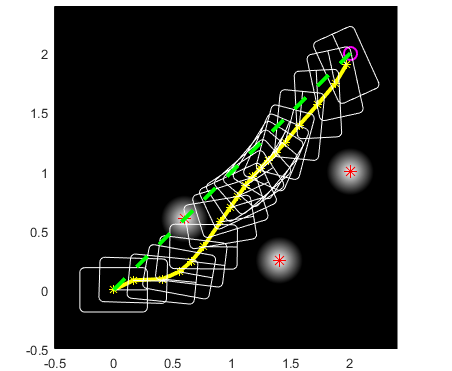}}\\
   \subfloat[Range-squared]{\includegraphics[width=0.32\linewidth]{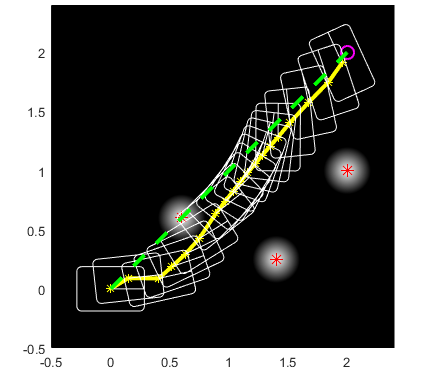}}
   \subfloat[Light-dark \cite{Platt10}]{\includegraphics[width=0.32\linewidth]{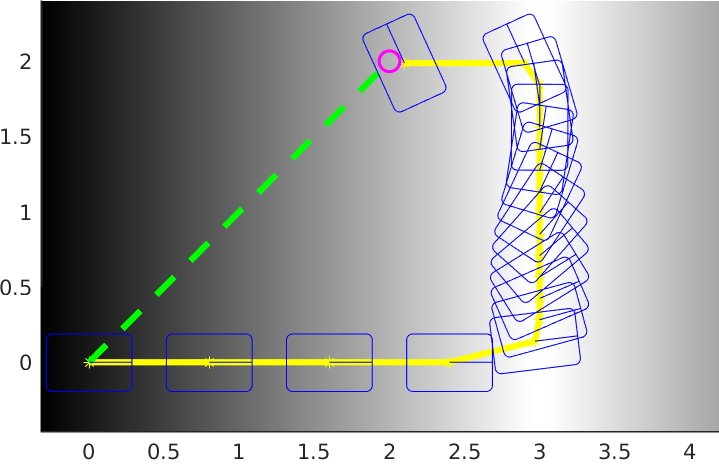}}
   \subfloat[Light-dark \cite{platt2013convex}]{\includegraphics[width=0.32\linewidth]{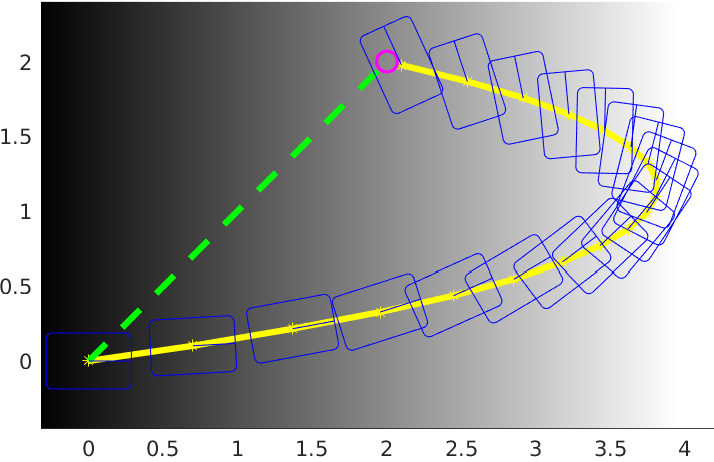}}
   \caption{Simulation results for obstacle-free situation with different observation models. The information is color-coded. Lighter means less noisy observations. The dashed green line represents the initial trajectory; the solid yellow line shows the optimized trajectory. In all cases, $ \hat{\mathbf{x}}_0=(0, 0, 0) $, $ \mathbf{x}_g=(2, 2, 2) $, and $ r_g=0.1 $.}
  \label{fig:obstacle-free cases} 
\end{figure}
\begin{figure*}[t!]
\centering
   \subfloat[]{\includegraphics[width=0.5\linewidth]{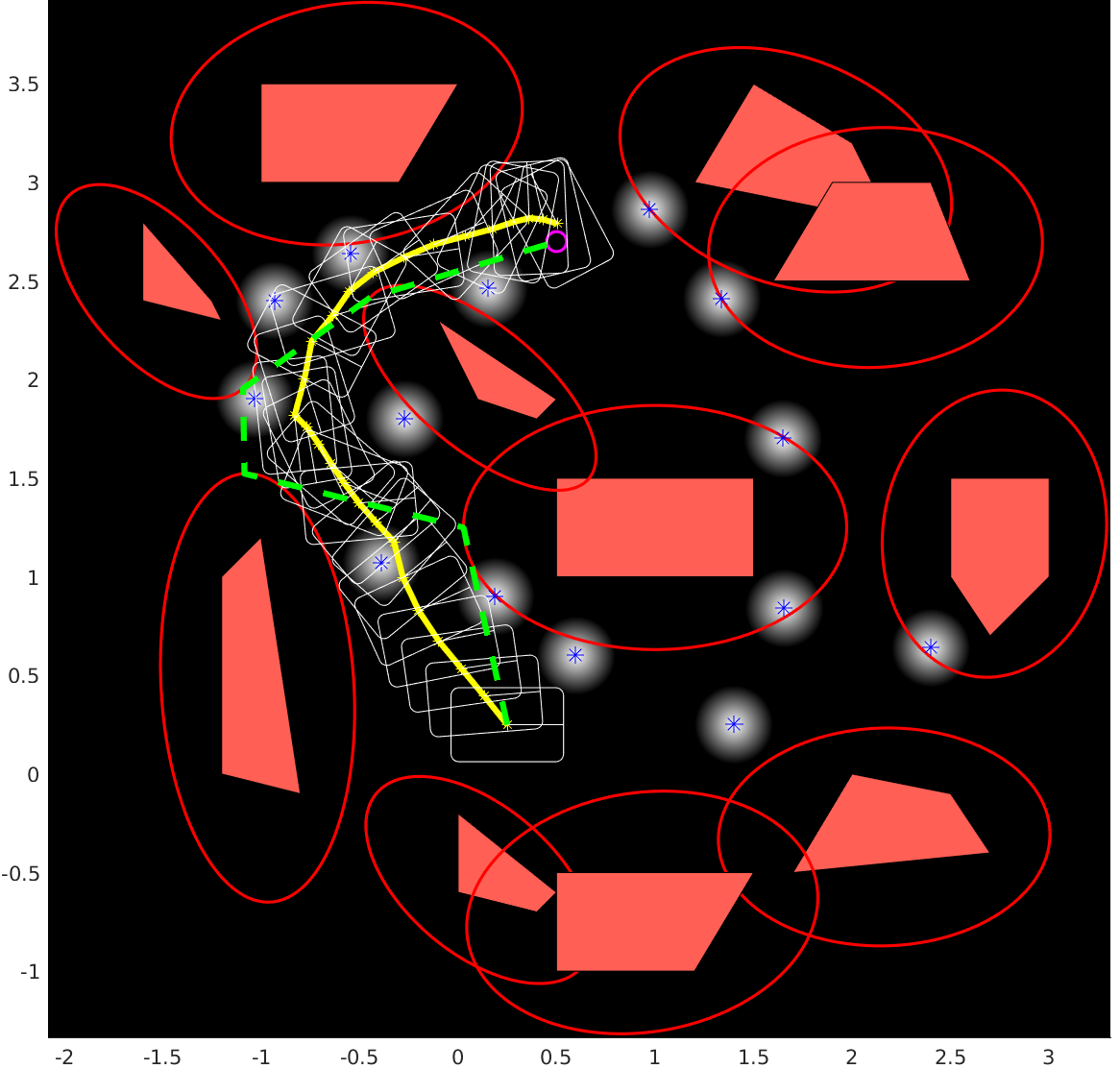}}
   \subfloat[]{\includegraphics[width=0.5\linewidth]{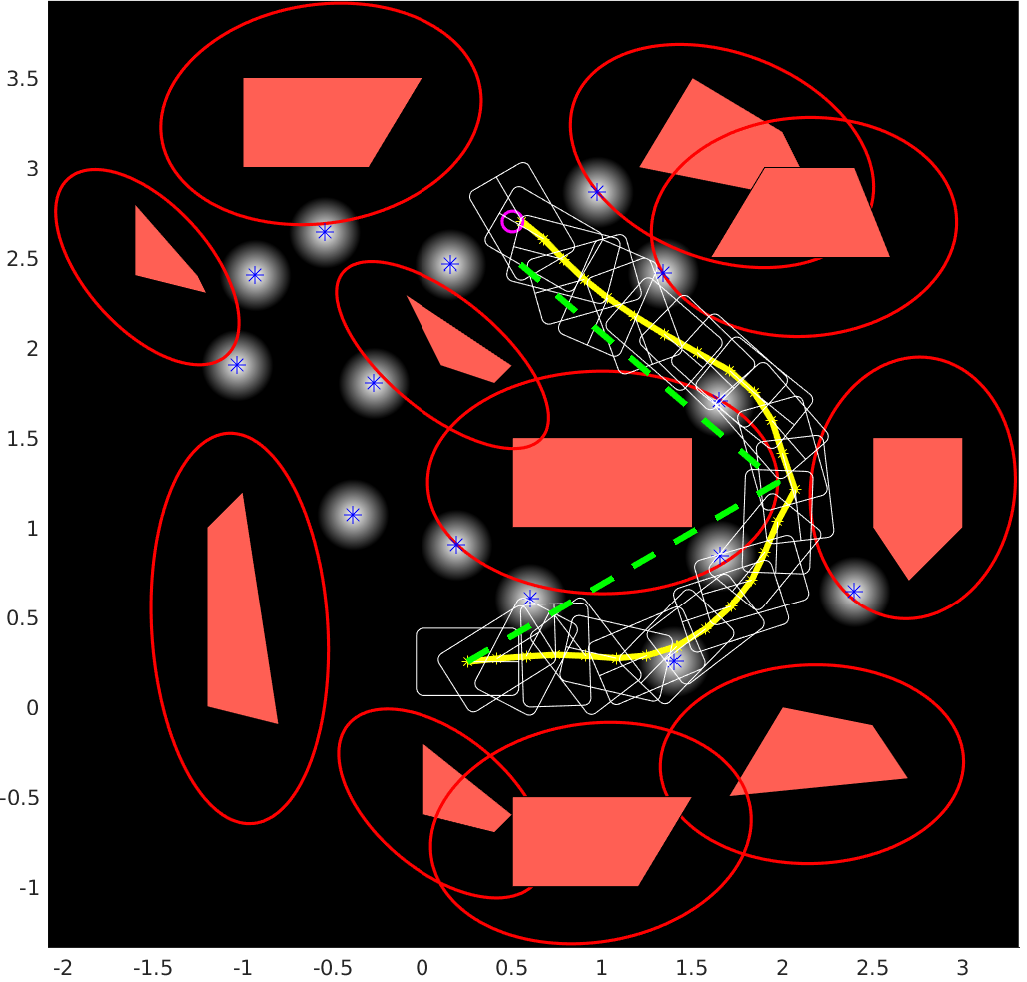}\label{fig:obstacle cases, b}}
   \caption{Simulation results for two different initializations with obstacles. The obstacles are the red solid polygons; the ellipses show the inflated regions around them avoided by the configuration of points that represent the robot (they are also the argument of the Gaussian function in the obstacle cost). In all cases, $ \hat{\mathbf{x}}_0=(0.25, 0.25, 0) $, $ \mathbf{x}_g=(0.5, 2.7, 2) $, and $ r_g=0.1 $. The optimized trajectory in case (b) has a lower overall cost.}
  \label{fig:obstacle cases} 
\end{figure*}
\section{Simulation Results}\label{sec:Simulation Results}
In this section, we provide our simulation results to show the performance of T-LQG. Our simulations are performed in MATLAB 2016a with a 2.90 GHz CORE i7 machine with dual core technology and 8 GB of RAM. We use the MATLAB's fmincon solver to solve the NLP problem. First, we provide the overall algorithm and the overall control loop. Then, we investigate several situations in which the environment is obstacle-free. We perform 9 simulations for a KUKA youBot base model, with 9 different observation models including models adapted from the literature. Then, we perform a simulation in a complex environment with many obstacles. We conduct this scenario for two different initial trajectories and compare the results. In each scenario, we show the initial trajectory used to initialize the optimization problem along with the optimized output trajectory.

\textit{Implementation:} The overall control loop is shown in Fig. \ref{fig:control_Loop}, and the overall T-LQG algorithm is reflected in Algorithm \ref{alg:re-planning}. As it is seen in Fig. \ref{fig:control_Loop} and Alg. \ref{alg:re-planning}, the planning problem starts with the supply of an initial belief and ends whenever the probability of reaching the goal region is greater than a predefined threshold $ p_g>0 $. The planner $ \pi $ is fed the initial belief $ \mathbf{b}_0 $, the obstacle parameters $ (\mathcal{E},\mathcal{C}) $, planning horizon $ K $, a goal state $ \mathbf{x}_g $, the goal region radius $ r_g $, and other parameters, such as system equations. The resultant planned trajectory is provided to the controller, whose output is the policy function. The policy is executed, a new observation is perceived, and a new belief is obtained. If the distance between the updated belief and the nominal belief $ d(\mathbf{b}_t, \mathbf{b}^{o}_t)>d_{th} $ is greater than a threshold, or the policy execution is finished but the criteria is not met, the planning algorithm restarts.

\textit{Obstacle-free environment:} Let us use the kinematics equations of KUKA youBot base as described in \cite{1_zakharov_2011}. Particularly, the state vector can be denoted by a 3D vector, $ \mathbf{x} = [\mathbf{x}_{\mathtt{x}}, \mathbf{x}_{\mathtt{y}}, \mathbf{x}_{\theta} ]^T $, which describes the position and heading of the robot base, and $ \mathbf{x}\in SO(2) $. The control consists of the velocities of the four wheels. It can be shown that the discrete motion model can be written as $ \mathbf{x}_{t+1} = f(\mathbf{x}_{t}, \mathbf{u}_t, \boldsymbol{\omega}_t) = \mathbf{x}_t+\mathbf{B}\mathbf{u}_tdt+\mathbf{G}\boldsymbol{\omega}_t\sqrt{dt} $, where $ \mathbf{B} $ and $ \mathbf{G} $ are appropriate constant matrices whose elements depend on the dimensions of the robot as indicated in \cite{2_youbot_store_com_2016}, and $ dt $ is the time-discretization period. The results depicted in Fig. \ref{fig:obstacle-free cases} are for different observation models; including the range and bearing; bearing-only and range-only observation from one landmark in cases (a)-(c); and from three landmarks in cases (d)-(f), respectively. In case (g), the observation function is changed to the square of the range function. Finally, in cases (h) and (i), the light-dark models of the papers \cite{Platt10} and \cite{platt2013convex} are adopted. In both cases, the observation functions are linear, and the covariance of the observation noise is state dependent. It is a quadratic function with a minimum at 3 in case (h) and a hyperbolic function with a minimum at $ +\infty $ in (i). More details of these functions can be found in the corresponding papers. Finally as it is noticed in all cases, the optimization is initialized with the trivial straight-line, which is reflected in the figures with the dashed green line, whereas the optimal trajectories are depicted with solid lines.

\textit{Complex environment:} Next, we perform a simulation in an environment full of obstacles for the youBot, with range and bearing observations from several landmarks. Inspired by \cite{zucker2013chomp}, we model the robot with a configuration of a set of points that represent the balls' centers that cover the body of the robot. In our simulations, only two balls whose radii are proportional to the width of the robot suffice to cover the robot \cite{rafieisakhaei2016nonGSLAP}. We find the MVEE of the polygons  that are inflated from each vertex to the size of the radius and modify the cost of obstacles to keep the centers of the balls out of the new barriers. As it is seen in Fig. \ref{fig:obstacle cases}. we have initialed the optimization problem with two different initial trajectories obtained using a modified viability graph algorithm (shown with the green dashed lines). It should be noted that there is nothing particular about the initialization algorithm and methods---a planner such as RRT can be used as well, as long as the initialization trajectory is semi-feasible in that it does not pass through the infeasible local minima of the barrier functions. As it seen in this figure, the planning horizon is large (26 steps in case (a) and 25 steps in case (b)), which shows the scalability of T-LQG. The results show that the optimized trajectory (reflected with solid lines) avoids entering the banned regions bordered by the ellipsoids, so that the robot itself avoids colliding with the obstacles. Moreover, the locally optimal trajectory gets closer to the information sources and thereby obtains the best predicted estimation performance. In this scenario, by comparing the cost of the two optimize trajectories, the better of the two (trajectory in Fig. \ref{fig:obstacle cases, b}) is chosen as the plan for execution.

\section{Conclusion}\label{sec:Conclusion} 
In this paper, we simplified the solution of the belief space planning problem by proposing a scalable method that is backed by theoretical analysis supported by the control literature. Particularly, we proposed a deterministic optimal control problem that can be solved by an NLP solver with $ O(Kn^3) $ computational complexity.  The goal of Trajectory-optimized LQG is to find an LQG policy with the best nominal performance. T-LQG achieves this by finding the best underlying linearization trajectory for a non-linear system with a non-linear observation model, utilizing the trajectory-dependent covariance evolution of the Kalman filter given by the dynamic Riccati equations. We could do this by the proper usage of the separation principle that provided us with an LQR controller for a linearized system along that nominal trajectory. We proved that the accumulated error that is resulted by our calculations is deterministic under a first-order approximation and only depends on the linearization error. This can be overcome by either increasing the linearization points or by replanning whenever the deviation from the planned trajectory is higher than a predefined tolerance. We also extended the method to non-convex environments by adding a cost function to avoid collision with the obstacles. Finally, we performed simulations for a common robotic system with several observation functions in obstacle-free environments, as well as complex narrow passages with obstacles. 

In conclusion, while T-LQG and the MLO method of \cite{Platt10} address a similar optimization problem, their theoretical approach is vastly different:
\begin{itemize}\compresslist
\item[\textbullet]where MLO uses a heuristic approach, T-LQG uses the separation principle;  

\item[\textbullet]MLO does not have a controller in the design, whereas T-LQG does;

\item[\textbullet]MLO uses assumptions on the observations to reach the optimization problem, while in T-LQG, assumptions on observations are inconsequential;

\item[\textbullet]MLO designs a belief-LQR, but T-LQG only requires an LQR on the state;

\item[\textbullet]MLO starts with an EKF design and linearizes the system equations around the mean update, while T-LQG starts with linearizing the system equations around a nominal trajectory and uses the KF and separation principle to obtain the nominal performance around that trajectory; 
\item[\textbullet]MLO assumes from the beginning that process noise does not exist, and, ultimately, assumes observation noise does not exist either, but in T-LQG, neither of these assumptions exist; and

\item[\textbullet]while the computational complexity for MLO is $O(n_{tr}(Kn^3+kn^2))$ or $O(n_{tr}($  $Kn^3+kn^2)+ Kn^6)$, T-LQG minimizes the complexity to $ O(Kn^3) $.
\end{itemize}

Our future works will extend the theory and test the validity of our results for more complex situations.

\section*{Acknowledgment}
\addcontentsline{toc}{section}{Acknowledgment}
This material is based upon work partially supported by NSF under Contract Nos. CNS-1646449 and Science \& Technology Center Grant CCF-0939370, the U.S. Army Research Office under Contract No. W911NF-15-1-0279, and NPRP grant NPRP 8-1531-2-651 from the Qatar National Research Fund, a member of Qatar Foundation. 

\bibliographystyle{IEEEtran}

\bibliography{AliAgha}
\appendices
\section{Expressing the Cost Function as a Function of The Belief}\label{appdx:Expressing the Cost Function as a Function of The Belief}

In this appendix, we show that the cost function in \eqref{eq:Estimator cost}, expressed as a function of state and control, is indeed a function of belief and the control actions.

\textit{Estimation cost:} The optimization cost function, $ J $, can be rewritten as follows:
\begin{align*}
J=&\mathbb{E}[\sum_{t=1}^{K}\check{\mathbf{x}}^{T}_t\mathbf{W}^{x}_{t}\check{\mathbf{x}}_t+\tilde{\mathbf{u}}^{T}_{t-1}\mathbf{W}^{u}_{t}\tilde{\mathbf{u}}_{t-1}]
\\=&\mathbb{E}[\sum_{t=1}^{K}(\mathbf{x}_t-\hat{\mathbf{x}}_{t})^{T}\mathbf{W}^{x}_{t}(\mathbf{x}_t-\hat{\mathbf{x}}_{t})+\tilde{\mathbf{u}}^{T}_{t-1}\mathbf{W}^{u}_{t}\tilde{\mathbf{u}}_{t-1}]
\\=&\mathbb{E}[\sum_{t=1}^{K}(\mathbf{x}_t-\hat{\mathbf{x}}_{t})^{T}\mathbf{W}^{T}_{t}\mathbf{W}_{t}(\mathbf{x}_t-\hat{\mathbf{x}}_{t})+\tilde{\mathbf{u}}^{T}_{t-1}\mathbf{W}^{u}_{t}\tilde{\mathbf{u}}_{t-1}]
\\=&\mathbb{E}[\sum_{t=1}^{K}(\mathbf{W}_{t}(\mathbf{x}_t-\hat{\mathbf{x}}_{t}))^{T}(\mathbf{W}_{t}(\mathbf{x}_t-\hat{\mathbf{x}}_{t}))+\tilde{\mathbf{u}}^{T}_{t-1}\mathbf{W}^{u}_{t}\tilde{\mathbf{u}}_{t-1}]
\\=&\mathbb{E}[\sum_{t=1}^{K}\mathrm{tr}[(\mathbf{W}_{t}(\mathbf{x}_t-\hat{\mathbf{x}}_{t}))(\mathbf{W}_{t}(\mathbf{x}_t-\hat{\mathbf{x}}_{t}))^{T}]+\tilde{\mathbf{u}}^{T}_{t-1}\mathbf{W}^{u}_{t}\tilde{\mathbf{u}}_{t-1}]
\\=&\mathbb{E}[\sum_{t=1}^{K}\mathrm{tr}[\mathbf{W}_{t}(\mathbf{x}_t-\hat{\mathbf{x}}_{t})(\mathbf{x}_t-\hat{\mathbf{x}}_{t})^{T}\mathbf{W}^{T}_{t}]+\tilde{\mathbf{u}}^{T}_{t-1}\mathbf{W}^{u}_{t}\tilde{\mathbf{u}}_{t-1}]
\\=&\sum_{t=1}^{K}(\mathrm{tr}[\mathbf{W}_{t}\mathbb{E}[(\mathbf{x}_t-\hat{\mathbf{x}}_{t})(\mathbf{x}_t-\hat{\mathbf{x}}_{t})^{T}]\mathbf{W}^{T}_{t}]+\mathbb{E}[\tilde{\mathbf{u}}^{T}_{t-1}\mathbf{W}^{u}_{t}\tilde{\mathbf{u}}_{t-1}])
\\=&\sum_{t=1}^{K}(\mathrm{tr}[\mathbf{W}_{t}\mathbf{P}_{\mathbf{b}_t}\mathbf{W}^{T}_{t}]+\mathbb{E}[\tilde{\mathbf{u}}^{T}_{t-1}\mathbf{W}^{u}_{t}\tilde{\mathbf{u}}_{t-1}]),
\end{align*}
where we have used the fact that since $ \mathbf{W}^{x}_{t} $ is symmetric and positive semidefinite, there exists a (non-unique) Cholesky decomposition of $ \mathbf{W}^{x}_{t}=\mathbf{W}^{T}_{t}\mathbf{W}_{t} $, where the diagonal entries of the real upper triangular matrix $ \mathbf{W}_{t} $ can be zero \cite{watkins2004fundamentals}. Note, the Cholesky decomposition is unique, if and only if the $ \mathbf{W}^{x}_{t} $ is symmetric and positive definite. In such a case, the diagonal entries of $ \mathbf{W}_{t} $ are only positive. Moreover, $ \mathbf{P}_{\mathbf{b}_t}:=\mathbb{E}[(\mathbf{x}_t-\hat{\mathbf{x}}_{t})(\mathbf{x}_t-\hat{\mathbf{x}}_{t})^{T}] $ is the covariance of the belief during execution. Similarly, the nominal part of the cost function, $ J^P $, can also be written as a function of the nominal belief and the control actions using the definition of $ J^{p}=\sum_{t=0}^{K-1}c_t(\mathbf{b}^{p}_t,\mathbf{u}^{p}_t)+c_K(\mathbf{b}^{p}_K) $ in the section \ref{sec:Theoretical Considerations}, as follows:
\begin{align*}
J^{p}=&\sum_{t=1}^{K}(\mathbb{E}[\tilde{\mathbf{x}}^{T}_t\mathbf{W}^{x}_{t}\tilde{\mathbf{x}}_t]+({\mathbf{u}}^{p}_{t-1})^{T}\mathbf{W}^{u}_{t}{\mathbf{u}}^{p}_{t-1})
\\=&\sum_{t=1}^{K}(\mathbb{E}[(\mathbf{x}_t-\mathbf{x}^{p}_{t})^{T}\mathbf{W}^{x}_{t}(\mathbf{x}_t-\mathbf{x}^{p}_{t})]+({\mathbf{u}}^{p}_{t-1})^{T}\mathbf{W}^{u}_{t}{\mathbf{u}}^{p}_{t-1})
\\=&\sum_{t=1}^{K}(\mathrm{tr}[\mathbf{W}_{t}\mathbb{E}[(\mathbf{x}_t-\mathbf{x}^{p}_{t})(\mathbf{x}_t-\mathbf{x}^{p}_{t})^{T}]\mathbf{W}^{T}_{t}]+({\mathbf{u}}^{p}_{t-1})^{T}\mathbf{W}^{u}_{t}{\mathbf{u}}^{p}_{t-1})
\\=&\sum_{t=1}^{K}(\mathrm{tr}(\mathbf{W}_t \mathbf{P}^{+}_{\mathbf{b}^{p}_t}\mathbf{W}^T_t)+({\mathbf{u}}^{p}_{t-1})^{T}\mathbf{W}^{u}_{t}{\mathbf{u}}^{p}_{t-1}),
\end{align*}
where $ \mathbf{P}^{+}_{\mathbf{b}^{p}_t}=\mathbb{E}[(\mathbf{x}_t-\mathbf{x}^{p}_{t})(\mathbf{x}_t-\mathbf{x}^{p}_{t})^T] $ is the covariance of the nominal belief along the $ p\textendash traj $.

\section{Kalman Filtering Mean Update}\label{appdx:Kalman Filtering Mean Update}
In this appendix, we provide more details on the formula given in equation \eqref{eq:KF mean update}.

\textit{Mean update equations of KF:} Let us define $ \mathbf{e}_{\hat{\mathbf{x}}_{t}}:=\hat{\mathbf{x}}_{t}-\mathbf{x}^{p}_{t} $ and $ \mathbf{e}_{\hat{\mathbf{x}}^{-}_{t}}:=\hat{\mathbf{x}}^{-}_{t}-\mathbf{x}^{p}_{t} $ define predicted and updated errors of LQG estimation, respectively, where $ \hat{\mathbf{x}}^{-}_{t} $ is the predicted estimate of the system obtained in the prediction step of the KF. In a Kalman filter for a system with equation given in \eqref{eq:linearized system}, in the prediction step, $ \mathbf{e}_{\hat{\mathbf{x}}^{-}_{t+1}} $ is calculated as follows \cite{Ali14}:
\begin{align} \mathbf{e}_{\hat{\mathbf{x}}^{-}_{t+1}} = \mathbf{A}^{p}_{t}\mathbf{e}_{\hat{\mathbf{x}}_{t}}+\mathbf{B}^{p}_{t}(\mathbf{u}_{t}-\mathbf{u}^{p}_{t}).\label{eq:LQG error prediction}
\end{align}
Moreover, in the update step of the KF, $ \mathbf{e}_{\hat{\mathbf{x}}^{+}_{t+1}} $ is obtained as follows \cite{Ali14}:
\begin{align}
\mathbf{e}_{\hat{\mathbf{x}}_{t+1}}=\mathbf{e}_{\hat{\mathbf{x}}^{-}_{t+1}}+\mathbf{K}^{p}_{t+1}((\mathbf{z}_t-\mathbf{z}^{p}_{t})-\mathbf{H}^{p}_{t+1}\mathbf{e}_{\hat{\mathbf{x}}^{-}_{t+1}}),\label{eq:LQG error update}
\end{align}
where $ \mathbf{K}^{p}_{t+1} $ is the Kalman gain as defined in equation \eqref{eq:riccati 3}. Therefore, using equation \eqref{eq:LQG error prediction}, \eqref{eq:LQG error update} can be rewritten as follows:
\begin{align*}
\mathbf{e}_{\hat{\mathbf{x}}_{t+1}}=&\mathbf{A}^{p}_{t}\mathbf{e}_{\hat{\mathbf{x}}^{+}_{t}}+\mathbf{B}^{p}_{t}(\mathbf{u}_{t}-\mathbf{u}^{p}_{t})+\mathbf{K}^{p}_{t+1}((\mathbf{z}_t-\mathbf{z}^{p}_{t})-\mathbf{H}^{p}_{t+1}(\mathbf{A}^{p}_{t}\mathbf{e}_{\hat{\mathbf{x}}^{+}_{t}}+\mathbf{B}^{p}_{t}(\mathbf{u}_{t}-\mathbf{u}^{p}_{t})))
\\=&\mathbf{A}^{p}_{t}(\hat{\mathbf{x}}_{t}-\mathbf{x}^{p}_{t})+\mathbf{B}^{p}_{t}(\mathbf{u}_{t}-\mathbf{u}^{p}_{t})+\mathbf{K}^{p}_{t+1}((\mathbf{z}_t-\mathbf{z}^{p}_{t})-\mathbf{H}^{p}_{t+1}(\mathbf{A}^{p}_{t}(\hat{\mathbf{x}}_{t}-\mathbf{x}^{p}_{t})+\mathbf{B}^{p}_{t}(\mathbf{u}_{t}-\mathbf{u}^{p}_{t})))
\\=&\mathbf{x}^{p}_{t+1}-\mathbf{x}^{p}_{t+1}+\mathbf{A}^{p}_{t}(\hat{\mathbf{x}}_{t}-\mathbf{x}^{p}_{t})+\mathbf{B}^{p}_{t}(\mathbf{u}_{t}-\mathbf{u}^{p}_{t})
\\&+\mathbf{K}^{p}_{t+1}((\mathbf{z}_t-\mathbf{z}^{p}_{t})-\mathbf{H}^{p}_{t+1}(\mathbf{x}^{p}_{t+1}-\mathbf{x}^{p}_{t+1}+\mathbf{A}^{p}_{t}(\hat{\mathbf{x}}_{t}-\mathbf{x}^{p}_{t})+\mathbf{B}^{p}_{t}(\mathbf{u}_{t}-\mathbf{u}^{p}_{t})))
\\=&-\mathbf{x}^{p}_{t+1}+\mathbf{x}^{p}_{t+1}-\mathbf{A}^{p}_{t}\mathbf{x}^{p}_{t}-\mathbf{B}^{p}_{t}\mathbf{u}^{p}_{t}+\mathbf{A}^{p}_{t}\hat{\mathbf{x}}_{t}+\mathbf{B}^{p}_{t}\mathbf{u}_{t}
\\&+\mathbf{K}^{p}_{t+1}(\mathbf{z}_t-\mathbf{z}^{p}_{t}-\mathbf{H}^{p}_{t+1}(-\mathbf{x}^{p}_{t+1}+\mathbf{x}^{p}_{t+1}-\mathbf{A}^{p}_{t}\mathbf{x}^{p}_{t}-\mathbf{B}^{p}_{t}\mathbf{u}^{p}_{t}+\mathbf{A}^{p}_{t}\hat{\mathbf{x}}_{t}+\mathbf{B}^{p}_{t}\mathbf{u}_{t}))
\\=&-\mathbf{x}^{p}_{t+1}+\mathbf{f}^{p}_{t}+\mathbf{A}^{p}_{t}\hat{\mathbf{x}}_{t}+\mathbf{B}^{p}_{t}\mathbf{u}_{t}
+\mathbf{K}^{p}_{t+1}(\mathbf{z}_t-(\mathbf{z}^{p}_{t}-\mathbf{H}^{p}_{t+1}\mathbf{x}^{p}_{t+1})-\mathbf{H}^{p}_{t+1}(\mathbf{f}^{p}_{t}+\mathbf{A}^{p}_{t}\hat{\mathbf{x}}_{t}+\mathbf{B}^{p}_{t}\mathbf{u}_{t}))
\\=&-\mathbf{x}^{p}_{t+1}+\mathbf{f}^{p}_{t}-\mathbf{K}^{p}_{t+1}\mathbf{H}^{p}_{t+1}\mathbf{f}^{p}_{t}+\mathbf{A}^{p}_{t}\hat{\mathbf{x}}_{t}+\mathbf{B}^{p}_{t}\mathbf{u}_{t}
+\mathbf{K}^{p}_{t+1}(\mathbf{z}_t-\mathbf{h}^{p}_{t}-\mathbf{H}^{p}_{t+1}(\mathbf{A}^{p}_{t}\hat{\mathbf{x}}_{t}+\mathbf{B}^{p}_{t}\mathbf{u}_{t}))
\\=&-\mathbf{x}^{p}_{t+1}+(\mathbf{I}-\mathbf{K}^{p}_{t+1}\mathbf{H}^{p}_{t+1})\mathbf{f}^{p}_{t}-\mathbf{K}^{p}_{t+1}\mathbf{h}^{p}_{t}+\mathbf{A}^{p}_{t}\hat{\mathbf{x}}_{t}+\mathbf{B}^{p}_{t}\mathbf{u}_{t}
+\mathbf{K}^{p}_{t+1}(\mathbf{z}_t-\mathbf{H}^{p}_{t+1}(\mathbf{A}^{p}_{t}\hat{\mathbf{x}}_{t}+\mathbf{B}^{p}_{t}\mathbf{u}_{t})),
\end{align*}
where $ \mathbf{f}^{p}_{t}:=f(\mathbf{x}^{p}_{t},\mathbf{u}^{p}_{t}, \mathbf{0})-\mathbf{A}^{p}_t\mathbf{x}^{p}_t-\mathbf{B}^{p}_t\mathbf{u}^{p}_t $ and $ \mathbf{h}^{p}_t:=h(\mathbf{x},\mathbf{0})- \mathbf{H}^{p}_t\mathbf{x}^{p}_t $. Therefore, using the definition of the $ \mathbf{e}_{\hat{\mathbf{x}}_{t+1}} $, the mean update equation for a linearized system around the $ o\textendash traj $ (which is a particularly a $ p\textendash traj $) can be calculated as follows:
\begin{align*}
\hat{\mathbf{x}}_{t+1} = (\mathbf{I}-\mathbf{K}^{o}_{t+1}\mathbf{H}^{o}_{t+1})\mathbf{f}^{o}_{t}-\mathbf{K}^{o}_{t+1}\mathbf{h}^{o}_{t+1}+\mathbf{A}^{o}_t\hat{\mathbf{x}}_t + \mathbf{B}^{o}_t\mathbf{u}_t + \mathbf{K}^{o}_{t+1}(\mathbf{z}_{t+1}-\mathbf{H}^{o}_{t+1}(\mathbf{A}^{o}_t\hat{\mathbf{x}}_t + \mathbf{B}^{o}_t\mathbf{u}_t))
\end{align*}
with $ \hat{\mathbf{x}}_{0} = \mathbb{E}(\mathbf{x}_0) $.
\section{Proofs of Lemmas and Theorems}\label{appdx:Proofs of Lemmas and Theorems}
In this appendix, we provide the proof of Lemmas \ref{lemma:Estimation},\ref{lemma:State}, \ref{lemma:Control}, \ref{lemma:Observation}, \ref{lemma:Belief}, and Theorem \ref{theroem:Cost}.
\begin{IEEEproof}\textup{\textbf{Lemma \ref{lemma:Estimation}, Estimation Error Propagation}}

Let us calculate $ \check{\mathbf{x}}_{t+1} $ for $ 0\le t $ as follows:
\begin{align*}
\check{\mathbf{x}}_{t+1}=&\mathbf{x}_{t+1}-\hat{\mathbf{x}}_{t+1}
\\=&(\mathbf{x}^{p}_{t+1}+\tilde{\mathbf{x}}_{t+1}-\mathbf{A}_t\hat{\mathbf{x}}_{t}-\mathbf{B}_t\mathbf{u}_t-\mathbf{K}_{t+1}(\mathbf{z}_{t+1}-\mathbf{H}_{t+1}(\mathbf{A}_t\hat{\mathbf{x}}_{t}+\mathbf{B}_t\mathbf{u}_t)))-((\mathbf{I}-\mathbf{K}_{t+1}\mathbf{H}_{t+1})\mathbf{f}^{p}_t-\mathbf{K}_{t+1}\mathbf{h}^{p}_{t+1})
\\=&\mathbf{x}^{p}_{t+1}+\tilde{\mathbf{x}}_{t+1}-\mathbf{A}_t\hat{\mathbf{x}}_{t}-\tilde{\mathbf{x}}_{t+1}+\mathbf{A}_t\tilde{\mathbf{x}}_t - \mathbf{B}_t\mathbf{u}^{p}_t +\mathbf{G}_t\boldsymbol{\omega}_t-((\mathbf{I}-\mathbf{K}_{t+1}\mathbf{H}_{t+1})\mathbf{f}^{p}_t-\mathbf{K}_{t+1}\mathbf{h}^{p}_{t+1})
\\&-\mathbf{K}_{t+1}(\mathbf{z}^{p}_{t+1}+\mathbf{H}_{t+1}\tilde{\mathbf{x}}_{t+1}-\mathbf{H}_{t+1}(\mathbf{A}_t\hat{\mathbf{x}}_{t}+\tilde{\mathbf{x}}_{t+1}-\mathbf{A}_t\tilde{\mathbf{x}}_t + \mathbf{B}_t\mathbf{u}^{p}_t -\mathbf{G}_t\boldsymbol{\omega}_t)+\mathbf{M}_{t+1}\boldsymbol{\nu}_{t+1})
\\=&\mathbf{x}^{p}_{t+1}-\mathbf{A}_t\hat{\mathbf{x}}_{t}+\mathbf{A}_t\tilde{\mathbf{x}}_t - \mathbf{B}_t\mathbf{u}^{p}_t +\mathbf{G}_t\boldsymbol{\omega}_t-((\mathbf{I}-\mathbf{K}_{t+1}\mathbf{H}_{t+1})\mathbf{f}^{p}_t-\mathbf{K}_{t+1}\mathbf{h}^{p}_{t+1})
\\&-\mathbf{K}_{t+1}(\mathbf{z}^{p}_{t+1}-\mathbf{H}_{t+1}(\mathbf{A}_t\hat{\mathbf{x}}_{t}-\mathbf{A}_t\tilde{\mathbf{x}}_t + \mathbf{B}_t\mathbf{u}^{p}_t -\mathbf{G}_t\boldsymbol{\omega}_t)+\mathbf{M}_{t+1}\boldsymbol{\nu}_{t+1})
\\=&\mathbf{A}_t(\mathbf{x}_t-\hat{\mathbf{x}}_{t})+\mathbf{x}^{p}_{t+1}-\mathbf{A}_t\mathbf{x}^{p}_t - \mathbf{B}_t\mathbf{u}^{p}_t +\mathbf{G}_t\boldsymbol{\omega}_t-((\mathbf{I}-\mathbf{K}_{t+1}\mathbf{H}_{t+1})\mathbf{f}^{p}_t-\mathbf{K}_{t+1}\mathbf{h}^{p}_{t+1})
\\&-\mathbf{K}_{t+1}(\mathbf{z}^{p}_{t+1}+\mathbf{H}_{t+1}(\mathbf{A}_t(\mathbf{x}_t-\hat{\mathbf{x}}_{t})-\mathbf{A}_t\mathbf{x}^{p}_t - \mathbf{B}_t\mathbf{u}^{p}_t +\mathbf{G}_t\boldsymbol{\omega}_t)+\mathbf{M}_{t+1}\boldsymbol{\nu}_{t+1})
\\=&\mathbf{A}_t\check{\mathbf{x}}_{t}+\mathbf{f}^{p}_t +\mathbf{G}_t\boldsymbol{\omega}_t
-\mathbf{K}_{t+1}(\mathbf{z}^{p}_{t+1}+\mathbf{H}_{t+1}(\mathbf{A}_t\check{\mathbf{x}}_{t}+\mathbf{f}^{p}_t-\mathbf{x}^{p}_{t+1} +\mathbf{G}_t\boldsymbol{\omega}_t)+\mathbf{M}_{t+1}\boldsymbol{\nu}_{t+1})
\\&-((\mathbf{I}-\mathbf{K}_{t+1}\mathbf{H}_{t+1})\mathbf{f}^{p}_t-\mathbf{K}_{t+1}\mathbf{h}^{p}_{t+1})
\\=&\mathbf{A}_t\check{\mathbf{x}}_{t}+\mathbf{f}^{p}_t +\mathbf{G}_t\boldsymbol{\omega}_t
-\mathbf{K}_{t+1}(\mathbf{h}^{p}_{t+1}+\mathbf{H}_{t+1}(\mathbf{A}_t\check{\mathbf{x}}_{t}+\mathbf{f}^{p}_t +\mathbf{G}_t\boldsymbol{\omega}_t)+\mathbf{M}_{t+1}\boldsymbol{\nu}_{t+1})-((\mathbf{I}-\mathbf{K}_{t+1}\mathbf{H}_{t+1})\mathbf{f}^{p}_t-\mathbf{K}_{t+1}\mathbf{h}^{p}_{t+1})
\\=&(\mathbf{I}-\mathbf{K}_{t+1}\mathbf{H}_{t+1})\mathbf{A}_t\check{\mathbf{x}}_{t}+(\mathbf{I}-\mathbf{K}_{t+1}\mathbf{H}_{t+1})\mathbf{G}_t\boldsymbol{\omega}_t
-\mathbf{K}_{t+1}\mathbf{M}_{t+1}\boldsymbol{\nu}_{t+1}
\\&+((\mathbf{I}-\mathbf{K}_{t+1}\mathbf{H}_{t+1})\mathbf{f}^{p}_t-\mathbf{K}_{t+1}\mathbf{h}^{p}_{t+1})-((\mathbf{I}-\mathbf{K}_{t+1}\mathbf{H}_{t+1})\mathbf{f}^{p}_t-\mathbf{K}_{t+1}\mathbf{h}^{p}_{t+1})
\\=&:\mathbf{U}_{t+1}\mathbf{A}_t\check{\mathbf{x}}_{t}+\mathbf{U}_{t+1}\mathbf{G}_t\boldsymbol{\omega}_t
-\mathbf{K}_{t+1}\mathbf{M}_{t+1}\boldsymbol{\nu}_{t+1}
\\=&:\mathbf{F}_{t}\check{\mathbf{x}}_{t}+\mathbf{U}_{t+1}\mathbf{G}_t\boldsymbol{\omega}_t
-\mathbf{K}_{t+1}\mathbf{M}_{t+1}\boldsymbol{\nu}_{t+1}
\\=&:\tilde{\mathbf{F}}_{0:t}\check{\mathbf{x}}_{0}+\sum\limits_{s=0}^{t}\tilde{\mathbf{F}}_{s+1:t}(\mathbf{U}_{s+1}\mathbf{G}_s\boldsymbol{\omega}_s
-\mathbf{K}_{s+1}\mathbf{M}_{s+1}\boldsymbol{\nu}_{s+1})
\\=&\tilde{\mathbf{F}}_{0:t}\tilde{\mathbf{x}}_{0}+\sum\limits_{s=0}^{t}\tilde{\mathbf{F}}_{s+1:t}(\mathbf{U}_{s+1}\mathbf{G}_s\boldsymbol{\omega}_s
-\mathbf{K}_{s+1}\mathbf{M}_{s+1}\boldsymbol{\nu}_{s+1}).
\end{align*}
Note $ \check{\mathbf{x}}_0=\mathbf{x}_0-\mathbf{x}^{p}_{0}=\tilde{\mathbf{x}}_0 $. Also note the last formula is also correct for $ t=-1 $; since, $ \tilde{\mathbf{F}}_{0:-1} = \mathbf{I} $ from the definition, and the summation is zero based on Assumption \ref{assumption 1}.
\end{IEEEproof}

\begin{IEEEproof}\textup{\textbf{Lemma \ref{lemma:State}, State Error Propagation}}

The calculations of $ \tilde{\mathbf{x}}_{t+1} $ for $ t\ge 1 $ are as follows:
\begin{align*}
\tilde{\mathbf{x}}_{t+1}=&\mathbf{A}_t\tilde{\mathbf{x}}_t + \mathbf{B}_t\tilde{\mathbf{u}}_t +\mathbf{G}_t\boldsymbol{\omega}_t
\\=& (\mathbf{A}_t - \mathbf{B}_t\mathbf{L}_t)\tilde{\mathbf{x}}_t +\mathbf{G}_t\boldsymbol{\omega}_t
+\mathbf{B}_t(\tilde{\mathbf{F}}^{\mathbf{x}_0}_{t}\tilde{\mathbf{x}}_{0}+\sum\limits_{s=0}^{t-1}(\tilde{\mathbf{F}}^{\boldsymbol{\omega}}_{s,t}\boldsymbol{\omega}_s-\tilde{\mathbf{F}}^{\boldsymbol{\nu}}_{s+1,t}\boldsymbol{\nu}_{s+1}))
\\=&: \mathbf{D}_t\tilde{\mathbf{x}}_t +\mathbf{G}_t\boldsymbol{\omega}_t
+\mathbf{B}_t(\tilde{\mathbf{F}}^{\mathbf{x}_0}_{t}\tilde{\mathbf{x}}_{0}+\sum\limits_{s=0}^{t-1}(\tilde{\mathbf{F}}^{\boldsymbol{\omega}}_{s,t}\boldsymbol{\omega}_s-\tilde{\mathbf{F}}^{\boldsymbol{\nu}}_{s+1,t}\boldsymbol{\nu}_{s+1}))
\\=&:\tilde{\mathbf{D}}_{0:t}\tilde{\mathbf{x}}_0+\sum\limits_{r=1}^{t}\tilde{\mathbf{D}}_{r+1:t}[\mathbf{G}_r\boldsymbol{\omega}_r+\mathbf{B}_r(\tilde{\mathbf{F}}^{\mathbf{x}_0}_{r}\tilde{\mathbf{x}}_{0}+\sum\limits_{s=0}^{r-1}(\tilde{\mathbf{F}}^{\boldsymbol{\omega}}_{s,r}\boldsymbol{\omega}_s-\tilde{\mathbf{F}}^{\boldsymbol{\nu}}_{s+1,r}\boldsymbol{\nu}_{s+1}))]
\\=&\tilde{\mathbf{D}}_{0:t}\tilde{\mathbf{x}}_0+\sum\limits_{r=1}^{t}\tilde{\mathbf{D}}_{r+1:t}\mathbf{G}_r\boldsymbol{\omega}_r+\sum\limits_{r=1}^{t}\tilde{\mathbf{D}}_{r+1:t}\mathbf{B}_r\tilde{\mathbf{F}}^{\mathbf{x}_0}_{r}\tilde{\mathbf{x}}_{0}
+\sum\limits_{r=1}^{t}\sum\limits_{s=0}^{r-1}(\tilde{\mathbf{D}}_{r+1:t}\mathbf{B}_r\tilde{\mathbf{F}}^{\boldsymbol{\omega}}_{s,r}\boldsymbol{\omega}_s-\tilde{\mathbf{D}}_{r+1:t}\mathbf{B}_r\tilde{\mathbf{F}}^{\boldsymbol{\nu}}_{s+1,r}\boldsymbol{\nu}_{s+1})
\\=&:(\tilde{\mathbf{D}}_{0:t}+\sum\limits_{r=1}^{t}\tilde{\mathbf{D}}_{r+1:t}\mathbf{B}_r\tilde{\mathbf{F}}^{\mathbf{x}_0}_{r})\tilde{\mathbf{x}}_0+\sum\limits_{r=1}^{t}\tilde{\mathbf{D}}_{r+1:t}\mathbf{G}_r\boldsymbol{\omega}_r
+\sum\limits_{r=1}^{t}\sum\limits_{s=0}^{r-1}(\tilde{\mathbf{F}}^{\boldsymbol{\omega}}_{s,r,t}\boldsymbol{\omega}_s-\tilde{\mathbf{F}}^{\boldsymbol{\nu}}_{s+1,r,t}\boldsymbol{\nu}_{s+1})
\\=&:\tilde{\mathbf{D}}^{\mathbf{x}_{0}}_{t}\tilde{\mathbf{x}}_0+\sum\limits_{s=1}^{t}\tilde{\mathbf{D}}_{s+1:t}\mathbf{G}_s\boldsymbol{\omega}_s
+\sum\limits_{s=0}^{t-1}[(\sum\limits_{r=s+1}^{t}\tilde{\mathbf{F}}^{\boldsymbol{\omega}}_{s,r,t})\boldsymbol{\omega}_s-(\sum\limits_{r=s+1}^{t}\tilde{\mathbf{F}}^{\boldsymbol{\nu}}_{s+1,r,t})\boldsymbol{\nu}_{s+1}]
\\=&\tilde{\mathbf{D}}^{\mathbf{x}_{0}}_{t}\tilde{\mathbf{x}}_0+\sum\limits_{s=1}^{t}\tilde{\mathbf{D}}_{s+1:t}\mathbf{G}_s\boldsymbol{\omega}_s+\sum\limits_{s=0}^{t-1}(\sum\limits_{r=s+1}^{t}\tilde{\mathbf{F}}^{\boldsymbol{\omega}}_{s,r,t})\boldsymbol{\omega}_s
-\sum\limits_{s=0}^{t-1}(\sum\limits_{r=s+1}^{t}\tilde{\mathbf{F}}^{\boldsymbol{\nu}}_{s+1,r,t})\boldsymbol{\nu}_{s+1}
\\=&:\tilde{\mathbf{D}}^{\mathbf{x}_{0}}_{t}\tilde{\mathbf{x}}_0+\sum\limits_{s=0}^{t}\tilde{\mathbf{D}}^{\boldsymbol{\omega}}_{s,t}\boldsymbol{\omega}_s
-\sum\limits_{s=0}^{t-1}\tilde{\mathbf{D}}^{\boldsymbol{\nu}}_{s+1,t}\boldsymbol{\nu}_{s+1}.
\end{align*}
Note using the definition of $ \tilde{\mathbf{x}}_{t} $, the initial state error is $ \tilde{\mathbf{x}}_{0} = \mathbf{x}_0-\mathbf{x}^{p}_0 $. Likewise, the state error $ \tilde{\mathbf{x}}_{1} $ is $ \tilde{\mathbf{x}}_{1}=\mathbf{A}_0\tilde{\mathbf{x}}_0 +\mathbf{G}_0\boldsymbol{\omega}_0 $. Moreover, these errors are consistent with the formula given in the lemma using the definitions provided and Assumption \ref{assumption 1}.
\end{IEEEproof}

\begin{IEEEproof}\textup{\textbf{Lemma \ref{lemma:Control}, Control Error Propagation}}

\textit{Replacing estimation error in the control law:} Using equation \eqref{eq:Estimation Error Propagation}, we can rewrite $ \tilde{\mathbf{u}}_{t+1} $ for $ t\ge -1 $ as follows:
\begin{align}
\nonumber\tilde{\mathbf{u}}_{t+1}=&-\mathbf{L}_{t+1}(\hat{\mathbf{x}}_{t+1}-\mathbf{x}^{p}_{t+1})
\\\nonumber=&-\mathbf{L}_{t+1}(\hat{\mathbf{x}}_{t+1}-\mathbf{x}_{t+1}+\mathbf{x}_{t+1}-\mathbf{x}^{p}_{t+1})
\\\nonumber=&-\mathbf{L}_{t+1}\tilde{\mathbf{x}}_{t+1}-\mathbf{L}_{t+1}(\hat{\mathbf{x}}_{t+1}-\mathbf{x}_{t+1})
\\\nonumber=&-\mathbf{L}_{t+1}\tilde{\mathbf{x}}_{t+1}+\mathbf{L}_{t+1}\check{\mathbf{x}}_{t+1}
\\\nonumber=&-\mathbf{L}_{t+1}\tilde{\mathbf{x}}_{t+1}
+\mathbf{L}_{t+1}(\tilde{\mathbf{F}}_{0:t}\tilde{\mathbf{x}}_{0}+\sum\limits_{s=0}^{t}\tilde{\mathbf{F}}_{s+1:t}(\mathbf{U}_{s+1}\mathbf{G}_s\boldsymbol{\omega}_s-\mathbf{K}_{s+1}\mathbf{M}_{s+1}\boldsymbol{\nu}_{s+1}))
\\=:&-\mathbf{L}_{t+1}\tilde{\mathbf{x}}_{t+1}+\tilde{\mathbf{F}}^{\mathbf{x}_0}_{t+1}\tilde{\mathbf{x}}_{0}+\sum\limits_{s=0}^{t}(\tilde{\mathbf{F}}^{\boldsymbol{\omega}}_{s,t+1}\boldsymbol{\omega}_s-\tilde{\mathbf{F}}^{\boldsymbol{\nu}}_{s+1,t+1}\boldsymbol{\nu}_{s+1}).\label{eq:Replacing estimation error in the control law}
\end{align}

Now, let us simplify the control error propagation using equations \eqref{eq:Replacing estimation error in the control law} and \eqref{eq:State Error Propagation}. Thus, we can write $ \tilde{\mathbf{u}}_{t+1} $ for $ t\ge -1 $ as follows:
\begin{align*}
\tilde{\mathbf{u}}_{t+1}=&-\mathbf{L}_{t+1}\tilde{\mathbf{x}}_{t+1}+\tilde{\mathbf{F}}^{\mathbf{x}_0}_{t+1}\tilde{\mathbf{x}}_{0}+\sum\limits_{s=0}^{t}(\tilde{\mathbf{F}}^{\boldsymbol{\omega}}_{s,t+1}\boldsymbol{\omega}_s-\tilde{\mathbf{F}}^{\boldsymbol{\nu}}_{s+1,t+1}\boldsymbol{\nu}_{s+1})
\\=&-\mathbf{L}_{t+1}(\tilde{\mathbf{D}}^{\mathbf{x}_{0}}_{t}\tilde{\mathbf{x}}_0+\sum\limits_{s=0}^{t}\tilde{\mathbf{D}}^{\boldsymbol{\omega}}_{s,t}\boldsymbol{\omega}_s
-\sum\limits_{s=0}^{t-1}\tilde{\mathbf{D}}^{\boldsymbol{\nu}}_{s+1,t}\boldsymbol{\nu}_{s+1})+\tilde{\mathbf{F}}^{\mathbf{x}_0}_{t+1}\tilde{\mathbf{x}}_{0}+\sum\limits_{s=0}^{t}(\tilde{\mathbf{F}}^{\boldsymbol{\omega}}_{s,t+1}\boldsymbol{\omega}_s-\tilde{\mathbf{F}}^{\boldsymbol{\nu}}_{s+1,t+1}\boldsymbol{\nu}_{s+1})
\\=&-(\mathbf{L}_{t+1}\tilde{\mathbf{D}}^{\mathbf{x}_{0}}_{t}-\tilde{\mathbf{F}}^{\mathbf{x}_0}_{t+1})\tilde{\mathbf{x}}_0
-\sum\limits_{s=0}^{t}(\mathbf{L}_{t+1}\tilde{\mathbf{D}}^{\boldsymbol{\omega}}_{s,t}-\tilde{\mathbf{F}}^{\boldsymbol{\omega}}_{s,t+1})\boldsymbol{\omega}_s
-\sum\limits_{s=0}^{t-1}(\tilde{\mathbf{F}}^{\boldsymbol{\nu}}_{s+1,t+1}-\mathbf{L}_{t+1}\tilde{\mathbf{D}}^{\boldsymbol{\nu}}_{s+1,t})\boldsymbol{\nu}_{s+1}
-\tilde{\mathbf{F}}^{\boldsymbol{\nu}}_{t+1,t+1}\boldsymbol{\nu}_{t+1}
\\=&:-\mathbf{L}^{\mathbf{x}_{0}}_{t+1}\tilde{\mathbf{x}}_0
-\sum\limits_{s=0}^{t}\mathbf{L}^{\boldsymbol{\omega}}_{s,t+1}\boldsymbol{\omega}_s
-\sum\limits_{s=0}^{t}\mathbf{L}^{\boldsymbol{\nu}}_{s,t+1}\boldsymbol{\nu}_{s+1}.
\end{align*}
Note $ \tilde{\mathbf{u}}_{0}=\mathbf{0} $, and the last formula is consistent with this error using the definitions provided in the lemma and Assumption \ref{assumption 1}.
\end{IEEEproof}

\begin{IEEEproof}\textup{\textbf{Lemma \ref{lemma:Observation}, Observation Error Propagation}}

Let us calculate $ \tilde{\mathbf{z}}_{t+1} $ for $ t\ge 0 $ as follows:
\begin{align*}
\tilde{\mathbf{z}}_{t+1}=&\mathbf{H}_{t+1}\tilde{\mathbf{x}}_{t+1}+\mathbf{M}_{t+1}\boldsymbol{\nu}_{t+1}
\\=&\mathbf{H}_{t+1}(\tilde{\mathbf{D}}^{\mathbf{x}_{0}}_{t}\tilde{\mathbf{x}}_0+\sum\limits_{s=0}^{t}\tilde{\mathbf{D}}^{\boldsymbol{\omega}}_{s,t}\boldsymbol{\omega}_s
-\sum\limits_{s=0}^{t-1}\tilde{\mathbf{D}}^{\boldsymbol{\nu}}_{s+1,t}\boldsymbol{\nu}_{s+1})+\mathbf{M}_{t+1}\boldsymbol{\nu}_{t+1}
\\=&:\tilde{\mathbf{H}}^{\mathbf{x}_{0}}_{t+1}\tilde{\mathbf{x}}_0+\sum\limits_{s=0}^{t}\tilde{\mathbf{H}}^{\boldsymbol{\omega}}_{s,t+1}\boldsymbol{\omega}_s
+\sum\limits_{s=0}^{t}\tilde{\mathbf{H}}^{\boldsymbol{\nu}}_{s+1,t+1}\boldsymbol{\nu}_{s+1}.
\end{align*}
\end{IEEEproof}

\begin{IEEEproof}\textup{\textbf{Lemma \ref{lemma:Belief}, Belief Error Propagation}}

Utilizing the linearized equation of the  belief dynamics, we can obtain the propagated error of the belief, as well. First note that, since $ \tilde{\mathbf{b}}_{t+1}=\mathbf{T}^{\mathbf{b}}_t\tilde{\mathbf{b}}_t+\mathbf{T}^{\mathbf{u}}_t\tilde{\mathbf{u}}_t+ \mathbf{T}^{\mathbf{z}}_t\tilde{\mathbf{z}}_{t+1} $, we have $ \tilde{\mathbf{b}}_{t+1}=\sum\limits_{s=0}^{t}\tilde{\mathbf{T}}^{\mathbf{b}}_{s+1:t}(\mathbf{T}^{\mathbf{u}}_s\tilde{\mathbf{u}}_s+ \mathbf{T}^{\mathbf{z}}_s\tilde{\mathbf{z}}_{s+1}) $. Now, we can rewrite the belief error for $ t\ge 0 $ as follows:
\begin{align*}
\tilde{\mathbf{b}}_t=&\sum\limits_{s=0}^{t-1}\tilde{\mathbf{T}}^{\mathbf{b}}_{s+1:t-1}(\mathbf{T}^{\mathbf{u}}_s\tilde{\mathbf{u}}_s+ \mathbf{T}^{\mathbf{z}}_s\tilde{\mathbf{z}}_{s+1})
\\=&\tilde{\mathbf{T}}^{\mathbf{b}}_{1:t-1}\mathbf{T}^{\mathbf{u}}_0\tilde{\mathbf{u}}_0+\sum\limits_{s=1}^{t-1}\tilde{\mathbf{T}}^{\mathbf{b}}_{s+1:t-1}\mathbf{T}^{\mathbf{u}}_s\tilde{\mathbf{u}}_s+ \sum\limits_{s=0}^{t-1}\tilde{\mathbf{T}}^{\mathbf{b}}_{s+1:t-1}\mathbf{T}^{\mathbf{z}}_s\tilde{\mathbf{z}}_{s+1}
\\=&\sum\limits_{s=0}^{t-2}\tilde{\mathbf{T}}^{\mathbf{b}}_{s+2:t-1}\mathbf{T}^{\mathbf{u}}_{s+1}\tilde{\mathbf{u}}_{s+1}+ \sum\limits_{s=0}^{t-2}\tilde{\mathbf{T}}^{\mathbf{b}}_{s+1:t-1}\mathbf{T}^{\mathbf{z}}_s\tilde{\mathbf{z}}_{s+1}+\tilde{\mathbf{T}}^{\mathbf{b}}_{t:t-1}\mathbf{T}^{\mathbf{z}}_{t-1}\tilde{\mathbf{z}}_{t}
\\=&\sum\limits_{s=0}^{t-2}\tilde{\mathbf{T}}^{\mathbf{b}}_{s+2:t-1}\mathbf{T}^{\mathbf{u}}_{s+1}(-\mathbf{L}^{\mathbf{x}_{0}}_{s+1}\tilde{\mathbf{x}}_0
-\sum\limits_{r=0}^{s}\mathbf{L}^{\boldsymbol{\omega}}_{r,s+1}\boldsymbol{\omega}_r
-\sum\limits_{r=0}^{s}\mathbf{L}^{\boldsymbol{\nu}}_{r,s+1}\boldsymbol{\nu}_{r+1})
\\&+ \sum\limits_{s=0}^{t-2}\tilde{\mathbf{T}}^{\mathbf{b}}_{s+1:t-1}\mathbf{T}^{\mathbf{z}}_s(\tilde{\mathbf{H}}^{\mathbf{x}_{0}}_{s+1}\tilde{\mathbf{x}}_0+\sum\limits_{r=0}^{s}\tilde{\mathbf{H}}^{\boldsymbol{\omega}}_{r,s+1}\boldsymbol{\omega}_r
+\sum\limits_{r=0}^{s}\tilde{\mathbf{H}}^{\boldsymbol{\nu}}_{r+1,s+1}\boldsymbol{\nu}_{r+1})
\\&+\mathbf{T}^{\mathbf{z}}_{t-1}(\tilde{\mathbf{H}}^{\mathbf{x}_{0}}_{t}\tilde{\mathbf{x}}_0+\sum\limits_{s=0}^{t-1}\tilde{\mathbf{H}}^{\boldsymbol{\omega}}_{s,t}\boldsymbol{\omega}_s
+\sum\limits_{s=0}^{t-1}\tilde{\mathbf{H}}^{\boldsymbol{\nu}}_{s+1,t}\boldsymbol{\nu}_{s+1})
\\=&\sum\limits_{s=0}^{t-2}(-\tilde{\mathbf{T}}^{\mathbf{b}}_{s+2:t-1}\mathbf{T}^{\mathbf{u}}_{s+1}\mathbf{L}^{\mathbf{x}_{0}}_{s+1})\tilde{\mathbf{x}}_0
+\sum\limits_{s=0}^{t-2}\sum\limits_{r=0}^{s}(-\tilde{\mathbf{T}}^{\mathbf{b}}_{s+2:t-1}\mathbf{T}^{\mathbf{u}}_{s+1}\mathbf{L}^{\boldsymbol{\omega}}_{r,s+1})\boldsymbol{\omega}_r
+\sum\limits_{s=0}^{t-2}\sum\limits_{r=0}^{s}(-\tilde{\mathbf{T}}^{\mathbf{b}}_{s+2:t-1}\mathbf{T}^{\mathbf{u}}_{s+1}\mathbf{L}^{\boldsymbol{\nu}}_{r,s+1})\boldsymbol{\nu}_{r+1}
\\&+ \sum\limits_{s=0}^{t-2}(\tilde{\mathbf{T}}^{\mathbf{b}}_{s+1:t-1}\mathbf{T}^{\mathbf{z}}_s\tilde{\mathbf{H}}^{\mathbf{x}_{0}}_{s+1})\tilde{\mathbf{x}}_0
+\sum\limits_{s=0}^{t-2}\sum\limits_{r=0}^{s}(\tilde{\mathbf{T}}^{\mathbf{b}}_{s+1:t-1}\mathbf{T}^{\mathbf{z}}_s\tilde{\mathbf{H}}^{\boldsymbol{\omega}}_{r,s+1})\boldsymbol{\omega}_r
+\sum\limits_{s=0}^{t-2}\sum\limits_{r=0}^{s}(\tilde{\mathbf{T}}^{\mathbf{b}}_{s+1:t-1}\mathbf{T}^{\mathbf{z}}_s\tilde{\mathbf{H}}^{\boldsymbol{\nu}}_{r+1,s+1})\boldsymbol{\nu}_{r+1}
\\&+\mathbf{T}^{\mathbf{z}}_{t-1}\tilde{\mathbf{H}}^{\mathbf{x}_{0}}_{t}\tilde{\mathbf{x}}_0+\sum\limits_{s=0}^{t-1}\mathbf{T}^{\mathbf{z}}_{t-1}\tilde{\mathbf{H}}^{\boldsymbol{\omega}}_{s,t}\boldsymbol{\omega}_s
+\sum\limits_{s=0}^{t-1}\mathbf{T}^{\mathbf{z}}_{t-1}\tilde{\mathbf{H}}^{\boldsymbol{\nu}}_{s+1,t}\boldsymbol{\nu}_{s+1}
\\=&\sum\limits_{s=0}^{t-2}(-\tilde{\mathbf{T}}^{\mathbf{b}}_{s+2:t-1}\mathbf{T}^{\mathbf{u}}_{s+1}\mathbf{L}^{\mathbf{x}_{0}}_{s+1}+\tilde{\mathbf{T}}^{\mathbf{b}}_{s+1:t-1}\mathbf{T}^{\mathbf{z}}_s\tilde{\mathbf{H}}^{\mathbf{x}_{0}}_{s+1})\tilde{\mathbf{x}}_0+\mathbf{T}^{\mathbf{z}}_{t-1}\tilde{\mathbf{H}}^{\mathbf{x}_{0}}_{t}\tilde{\mathbf{x}}_0
\\
&+\sum\limits_{s=0}^{t-2}\sum\limits_{r=0}^{s}(-\tilde{\mathbf{T}}^{\mathbf{b}}_{s+2:t-1}\mathbf{T}^{\mathbf{u}}_{s+1}\mathbf{L}^{\boldsymbol{\omega}}_{r,s+1}+\tilde{\mathbf{T}}^{\mathbf{b}}_{s+1:t-1}\mathbf{T}^{\mathbf{z}}_s\tilde{\mathbf{H}}^{\boldsymbol{\omega}}_{r,s+1})\boldsymbol{\omega}_r
+\sum\limits_{s=0}^{t-1}\mathbf{T}^{\mathbf{z}}_{t-1}\tilde{\mathbf{H}}^{\boldsymbol{\omega}}_{s,t}\boldsymbol{\omega}_s
\\&+\sum\limits_{s=0}^{t-2}\sum\limits_{r=0}^{s}(-\tilde{\mathbf{T}}^{\mathbf{b}}_{s+2:t-1}\mathbf{T}^{\mathbf{u}}_{s+1}\mathbf{L}^{\boldsymbol{\nu}}_{r,s+1}+\tilde{\mathbf{T}}^{\mathbf{b}}_{s+1:t-1}\mathbf{T}^{\mathbf{z}}_s\tilde{\mathbf{H}}^{\boldsymbol{\nu}}_{r+1,s+1})\boldsymbol{\nu}_{r+1}
+\sum\limits_{s=0}^{t-1}\mathbf{T}^{\mathbf{z}}_{t-1}\tilde{\mathbf{H}}^{\boldsymbol{\nu}}_{s+1,t}\boldsymbol{\nu}_{s+1}
\\=&[\mathbf{T}^{\mathbf{z}}_{t-1}\tilde{\mathbf{H}}^{\mathbf{x}_{0}}_{t}+\sum\limits_{s=0}^{t-2}(-\tilde{\mathbf{T}}^{\mathbf{b}}_{s+2:t-1}\mathbf{T}^{\mathbf{u}}_{s+1}\mathbf{L}^{\mathbf{x}_{0}}_{s+1}+\tilde{\mathbf{T}}^{\mathbf{b}}_{s+1:t-1}\mathbf{T}^{\mathbf{z}}_s\tilde{\mathbf{H}}^{\mathbf{x}_{0}}_{s+1})]\tilde{\mathbf{x}}_0
\\
&+\sum\limits_{r=0}^{t-2}\sum\limits_{s=r}^{t-2}(-\tilde{\mathbf{T}}^{\mathbf{b}}_{s+2:t-1}\mathbf{T}^{\mathbf{u}}_{s+1}\mathbf{L}^{\boldsymbol{\omega}}_{r,s+1}+\tilde{\mathbf{T}}^{\mathbf{b}}_{s+1:t-1}\mathbf{T}^{\mathbf{z}}_s\tilde{\mathbf{H}}^{\boldsymbol{\omega}}_{r,s+1})\boldsymbol{\omega}_r
+\sum\limits_{s=0}^{t-1}\mathbf{T}^{\mathbf{z}}_{t-1}\tilde{\mathbf{H}}^{\boldsymbol{\omega}}_{s,t}\boldsymbol{\omega}_s
\\&+\sum\limits_{r=0}^{t-2}\sum\limits_{s=r}^{t-2}(-\tilde{\mathbf{T}}^{\mathbf{b}}_{s+2:t-1}\mathbf{T}^{\mathbf{u}}_{s+1}\mathbf{L}^{\boldsymbol{\nu}}_{r,s+1}+\tilde{\mathbf{T}}^{\mathbf{b}}_{s+1:t-1}\mathbf{T}^{\mathbf{z}}_s\tilde{\mathbf{H}}^{\boldsymbol{\nu}}_{r+1,s+1})\boldsymbol{\nu}_{r+1}
+\sum\limits_{s=0}^{t-1}\mathbf{T}^{\mathbf{z}}_{t-1}\tilde{\mathbf{H}}^{\boldsymbol{\nu}}_{s+1,t}\boldsymbol{\nu}_{s+1}
\\=&:\tilde{\mathbf{T}}^{\mathbf{x}_{0}}_{t}\tilde{\mathbf{x}}_0
+\sum\limits_{s=0}^{t-2}\sum\limits_{r=s}^{t-2}(-\tilde{\mathbf{T}}^{\mathbf{b}}_{r+2:t-1}\mathbf{T}^{\mathbf{u}}_{r+1}\mathbf{L}^{\boldsymbol{\omega}}_{s,r+1}+\tilde{\mathbf{T}}^{\mathbf{b}}_{r+1:t-1}\mathbf{T}^{\mathbf{z}}_r\tilde{\mathbf{H}}^{\boldsymbol{\omega}}_{s,r+1})\boldsymbol{\omega}_s
+\sum\limits_{s=0}^{t-2}\mathbf{T}^{\mathbf{z}}_{t-1}\tilde{\mathbf{H}}^{\boldsymbol{\omega}}_{s,t}\boldsymbol{\omega}_s
+\mathbf{T}^{\mathbf{z}}_{t-1}\tilde{\mathbf{H}}^{\boldsymbol{\omega}}_{t-1,t}\boldsymbol{\omega}_{t-1}
\\&+\sum\limits_{s=0}^{t-2}\sum\limits_{r=s}^{t-2}(-\tilde{\mathbf{T}}^{\mathbf{b}}_{r+2:t-1}\mathbf{T}^{\mathbf{u}}_{r+1}\mathbf{L}^{\boldsymbol{\nu}}_{s,r+1}+\tilde{\mathbf{T}}^{\mathbf{b}}_{r+1:t-1}\mathbf{T}^{\mathbf{z}}_r\tilde{\mathbf{H}}^{\boldsymbol{\nu}}_{s+1,r+1})\boldsymbol{\nu}_{s+1}
+\sum\limits_{s=0}^{t-2}\mathbf{T}^{\mathbf{z}}_{t-1}\tilde{\mathbf{H}}^{\boldsymbol{\nu}}_{s+1,t}\boldsymbol{\nu}_{s+1}+\mathbf{T}^{\mathbf{z}}_{t-1}\tilde{\mathbf{H}}^{\boldsymbol{\nu}}_{t,t}\boldsymbol{\nu}_{t}
\\=&\tilde{\mathbf{T}}^{\mathbf{x}_{0}}_{t}\tilde{\mathbf{x}}_0
+\sum\limits_{s=0}^{t-2}[\mathbf{T}^{\mathbf{z}}_{t-1}\tilde{\mathbf{H}}^{\boldsymbol{\omega}}_{s,t}+\sum\limits_{r=s}^{t-2}(-\tilde{\mathbf{T}}^{\mathbf{b}}_{r+2:t-1}\mathbf{T}^{\mathbf{u}}_{r+1}\mathbf{L}^{\boldsymbol{\omega}}_{s,r+1}+\tilde{\mathbf{T}}^{\mathbf{b}}_{r+1:t-1}\mathbf{T}^{\mathbf{z}}_r\tilde{\mathbf{H}}^{\boldsymbol{\omega}}_{s,r+1})]\boldsymbol{\omega}_s
+\mathbf{T}^{\mathbf{z}}_{t-1}\tilde{\mathbf{H}}^{\boldsymbol{\omega}}_{t-1,t}\boldsymbol{\omega}_{t-1}
\\&+\sum\limits_{s=0}^{t-2}[\mathbf{T}^{\mathbf{z}}_{t-1}\tilde{\mathbf{H}}^{\boldsymbol{\nu}}_{s+1,t}+\sum\limits_{r=s}^{t-2}(-\tilde{\mathbf{T}}^{\mathbf{b}}_{r+2:t-1}\mathbf{T}^{\mathbf{u}}_{r+1}\mathbf{L}^{\boldsymbol{\nu}}_{s,r+1}+\tilde{\mathbf{T}}^{\mathbf{b}}_{r+1:t-1}\mathbf{T}^{\mathbf{z}}_r\tilde{\mathbf{H}}^{\boldsymbol{\nu}}_{s+1,r+1})]\boldsymbol{\nu}_{s+1}
+\mathbf{T}^{\mathbf{z}}_{t-1}\tilde{\mathbf{H}}^{\boldsymbol{\nu}}_{t,t}\boldsymbol{\nu}_{t}
\\=&:\tilde{\mathbf{T}}^{\mathbf{x}_{0}}_{t}\tilde{\mathbf{x}}_0
+\sum\limits_{s=0}^{t-1}\tilde{\mathbf{T}}^{\boldsymbol{\omega}}_{s,t}\boldsymbol{\omega}_s
+\sum\limits_{s=1}^{t-1}[\mathbf{T}^{\mathbf{z}}_{t-1}\tilde{\mathbf{H}}^{\boldsymbol{\nu}}_{s,t}+\sum\limits_{r=s-1}^{t-2}(-\tilde{\mathbf{T}}^{\mathbf{b}}_{r+2:t-1}\mathbf{T}^{\mathbf{u}}_{r+1}\mathbf{L}^{\boldsymbol{\nu}}_{s-1,r+1}+\tilde{\mathbf{T}}^{\mathbf{b}}_{r+1:t-1}\mathbf{T}^{\mathbf{z}}_r\tilde{\mathbf{H}}^{\boldsymbol{\nu}}_{s,r+1})]\boldsymbol{\nu}_{s}
+\mathbf{T}^{\mathbf{z}}_{t-1}\tilde{\mathbf{H}}^{\boldsymbol{\nu}}_{t,t}\boldsymbol{\nu}_{t}
\\=&:\tilde{\mathbf{T}}^{\mathbf{x}_{0}}_{t}\tilde{\mathbf{x}}_0
+\sum\limits_{s=0}^{t-1}\tilde{\mathbf{T}}^{\boldsymbol{\omega}}_{s,t}\boldsymbol{\omega}_s
+\sum\limits_{s=1}^{t}\tilde{\mathbf{T}}^{\boldsymbol{\nu}}_{s,t}\boldsymbol{\nu}_{s}.
\end{align*}
Note $ \tilde{\mathbf{b}}_0=0 $, and this is consistent with the last formula, using the definitions given in the lemma and Assumption \ref{assumption 1}.
\end{IEEEproof}

\begin{IEEEproof}\textup{\textbf{Theorem \ref{theroem:Cost}, Cost Function Error}}

Using the linearization process described previously, we can write the cost function error as $ \tilde{J}= \mathbb{E}[\sum_{t=0}^{K-1} (\mathbf{C}^{\mathbf{b}}_t\tilde{\mathbf{b}}_t+ \mathbf{C}^{u}_t\tilde{\mathbf{u}}_t)+ \mathbf{C}^{\mathbf{b}}_K\tilde{\mathbf{b}}_K ] $. Utilizing the assumption that the process and observation noises are zero mean i.i.d., $ \mathbb{E}[\boldsymbol{\omega}_t]=\mathbb{E}[\boldsymbol{\nu}_t]={0} $ for all $ t $. Moreover, $ \mathbb{E}[\tilde{\mathbf{x}}_0]={0} $ which follows from the fact that $ \mathbb{E}[\mathbf{b}_0]=\mathbf{x}_0 $. Therefore, using the linearity of expectation operator and using Lemmas \ref{lemma:Control} and \ref{lemma:Belief}, we can rewrite $ \tilde{J} $ as follows:
\begin{align*}
\tilde{J}=&\sum_{t=0}^{K-1} (\mathbf{C}^{\mathbf{b}}_t\mathbb{E}[\tilde{\mathbf{b}}_t]+ \mathbf{C}^{u}_t\mathbb{E}[\tilde{\mathbf{u}}_t])+ \mathbf{C}^{\mathbf{b}}_K\mathbb{E}[\tilde{\mathbf{b}}_K]
\\=&\sum_{t=0}^{K} \mathbf{C}^{\mathbf{b}}_t\mathbb{E}[\tilde{\mathbf{T}}^{\mathbf{x}_{0}}_{t}\tilde{\mathbf{x}}_0
+\sum\limits_{s=0}^{t-1}\tilde{\mathbf{T}}^{\boldsymbol{\omega}}_{s,t}\boldsymbol{\omega}_s
+\sum\limits_{s=1}^{t}\tilde{\mathbf{T}}^{\boldsymbol{\nu}}_{s,t}\boldsymbol{\nu}_{s}]
+\sum_{t=0}^{K-1}\mathbf{C}^{u}_t\mathbb{E}[-\mathbf{L}^{\mathbf{x}_{0}}_{t+1}\tilde{\mathbf{x}}_0
-\sum\limits_{s=0}^{t}\mathbf{L}^{\boldsymbol{\omega}}_{s,t+1}\boldsymbol{\omega}_s
-\sum\limits_{s=0}^{t}\mathbf{L}^{\boldsymbol{\nu}}_{s,t+1}\boldsymbol{\nu}_{s+1}]
\\=&\sum_{t=0}^{K} \mathbf{C}^{\mathbf{b}}_t(\tilde{\mathbf{T}}^{\mathbf{x}_{0}}_{t}\mathbb{E}[\tilde{\mathbf{x}}_0]
+\sum\limits_{s=0}^{t-1}\tilde{\mathbf{T}}^{\boldsymbol{\omega}}_{s,t}\mathbb{E}[\boldsymbol{\omega}_s]
+\sum\limits_{s=1}^{t}\tilde{\mathbf{T}}^{\boldsymbol{\nu}}_{s,t}\mathbb{E}[\boldsymbol{\nu}_{s}])
+\sum_{t=0}^{K-1}\mathbf{C}^{u}_t(-\mathbf{L}^{\mathbf{x}_{0}}_{t+1}\mathbb{E}[\tilde{\mathbf{x}}_0]
-\sum\limits_{s=0}^{t}\mathbf{L}^{\boldsymbol{\omega}}_{s,t+1}\mathbb{E}[\boldsymbol{\omega}_s]
-\sum\limits_{s=0}^{t}\mathbf{L}^{\boldsymbol{\nu}}_{s,t+1}\mathbb{E}[\boldsymbol{\nu}_{s+1}])
\\=&\sum_{t=0}^{K} \mathbf{C}^{\mathbf{b}}_t(\tilde{\mathbf{T}}^{\mathbf{x}_{0}}_{t}\mathbf{0}
+\sum\limits_{s=0}^{t-1}\tilde{\mathbf{T}}^{\boldsymbol{\omega}}_{s,t}\mathbf{0}
+\sum\limits_{s=1}^{t}\tilde{\mathbf{T}}^{\boldsymbol{\nu}}_{s,t}\mathbf{0})
+\sum_{t=0}^{K-1}\mathbf{C}^{u}_t(-\mathbf{L}^{\mathbf{x}_{0}}_{t+1}\mathbf{0}
-\sum\limits_{s=0}^{t}\mathbf{L}^{\boldsymbol{\omega}}_{s,t+1}\mathbf{0}
-\sum\limits_{s=0}^{t}\mathbf{L}^{\boldsymbol{\nu}}_{s,t+1}\mathbf{0})
\\=&0.
\end{align*}
\end{IEEEproof}

\end{document}